\documentclass[10pt,twocolumn,letterpaper]{article}

\usepackage{iccv}              

\usepackage{amsthm}
\usepackage[table]{xcolor}
\usepackage{xcolor}
\usepackage{graphicx}
\usepackage{multirow}
\usepackage{tikz}

\usepackage[accsupp]{axessibility}  

\definecolor{iccvblue}{rgb}{0.21,0.49,0.74}
\usepackage[pagebackref,breaklinks,colorlinks,allcolors=iccvblue]{hyperref}

\def\paperID{14166}
\def\confName{ICCV}
\def\confYear{2025}

\title{{\em MosaicDiff}: Training-free Structural Pruning for Diffusion Model Acceleration Reflecting Pretraining Dynamics}

\author{Bowei Guo \quad Shengkun Tang \quad Cong Zeng \quad Zhiqiang Shen\\
Mohamed bin Zayed University of Artificial Intelligence\\
{\tt\small \{Bowei.Guo, Shengkun.Tang, Cong.Zeng, Zhiqiang.Shen\}@mbzuai.ac.ae}
}
\newtheorem{theorem}{Theorem}

\begin{document}

\twocolumn[{%
\renewcommand\twocolumn[1][]{#1}%

\maketitle
\vspace{-3em}
\begin{center}
    \centering
    \includegraphics[width=0.9\linewidth]{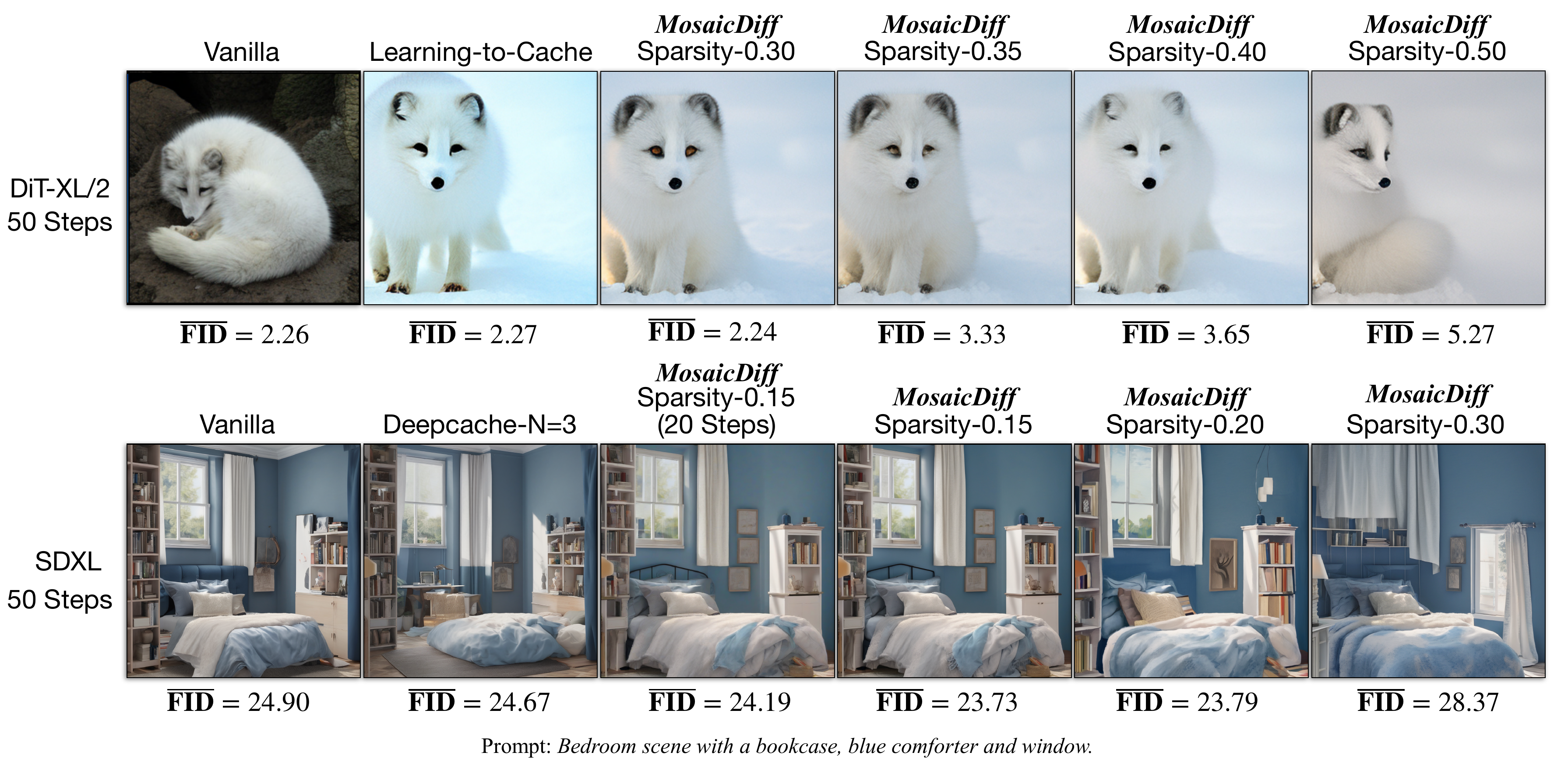}
    \vspace{-0.15in}
    \captionof{figure}{{\bf{\em MosaicDiff}} is a post-training / training-free structural pruning technique for both transformer-based and U-Net-based diffusion models. It can achieve 0.5 pruning sparsity on linear scheduled 675M DiT-XL/2 and 0.3 pruning sparsity on scaled-linear scheduled 2.6B SDXL-base-1.0 with minimal performance degradation.}
    \label{fig:visualization}
\end{center}

}]
\begin{abstract}

Diffusion models are renowned for their generative capabilities, yet their pretraining processes exhibit distinct phases of learning speed that have been entirely overlooked in prior post-training acceleration efforts in the community. In this study, we introduce a novel framework called {\bf{\em MosaicDiff}} that aligns diffusion pretraining dynamics with post-training sampling acceleration via trajectory-aware structural pruning. Our approach leverages the observation that the middle, fast-learning stage of diffusion pretraining requires more conservative pruning to preserve critical model features, while the early and later, slow-learning stages benefit from a more aggressive pruning strategy. This adaptive pruning mechanism is the first to explicitly mirror the inherent learning speed variations of diffusion pretraining, thereby harmonizing the model's inner training dynamics with its accelerated sampling process. Extensive experiments on DiT and SDXL demonstrate that our method achieves significant speed-ups in sampling without compromising output quality, outperforming previous state-of-the-art methods by large margins, also providing a new viewpoint for more efficient and robust training-free diffusion acceleration. Our implementation is available at \url{https://github.com/bwguo105/MosaicDiff.git}.

\end{abstract}    
\section{Introduction}
\label{sec:intro}

Diffusion models~\cite{ho2020denoising,rombach2022high,peebles2023scalable,ho2022classifier} have emerged as a powerful framework for generative tasks in unconditional image generation \cite{ho2020denoising}, text-guided image generation \cite{rombach2022high} and even 3D generation \cite{poole2022dreamfusion}, yet their extensive computational demands, especially during pretraining, pose significant challenges for real-world applications. The high cost of diffusion pretraining has driven the community toward training-free acceleration methods, which aim to reduce sampling times without incurring additional training overhead. However, these methods often overlook the nuanced learning dynamics inherent in the pretraining process.

A closer examination of diffusion pretraining reveals a unique characteristic: the learning speed is not uniform but varies significantly across different stages. In the middle phase, the model rapidly captures coarse-grained features, while the early and later stages involve a more gradual initial movement and final refinement of the details. This trajectory, marked by a {\em slow-fast-slow} multi-stage learning phase, provides crucial insights into how the diffusion model evolves over time, a factor that has been largely neglected in existing acceleration approaches.

The mainstream focus on training-free acceleration or pruning~\cite{lu2022dpm, lu2022dpm++, ma2024deepcache, tang2024adadiff, fang2023pruning, song2020denoising, luo2023latent}, driven by the desire to bypass the costly pretraining process, has unintentionally led to ignorance of the intrinsic learning properties of diffusion models. By not considering the differential learning speeds during pretraining, these methods miss an opportunity to optimize the post-training sampling process in a way that aligns with the model's internal learning dynamics. This oversight can limit the efficiency and effectiveness of the acceleration techniques employed.

To this end, we introduce a novel trajectory-aware pruning strategy called {\bf{\em MosaicDiff}} that aligns post-training / training-free acceleration with the underlying learning dynamics of the pretraining phase. By recognizing that the fast learning stage requires a more cautious approach while the slow learning stage can tolerate more aggressive pruning, our method strategically adjusts the pruning intensity along the model's learning trajectory. This alignment ensures that the post-training acceleration process is well-calibrated to the model's internal state, preserving critical features learned during the slower refinement stages. Specifically, we propose a new trajectory-aware second-order structural pruning method using SNR-aware (signal-to-noise ratio) calibration data, to identify {\em different sparse sub-networks from the same dense parent diffusion network} tailored to distinct learning stages. 

Even in our scenario where diffusion pretraining is not performed, our study demonstrates that it is possible to infer the learning speed characteristics from the sampling strategy itself. Through a combination of empirical studies and theoretical analysis, we connect the effective learning dynamics of the pretraining process to these insights, using them to guide our trajectory-aware pruning strategy. Our proposed methodology bridges the gap between pretraining behavior and post-training acceleration, allowing us to optimize the sampling process without the need for retraining.

Our experimental results show that this alignment strategy is particularly effective, especially at higher and more challenging pruning ratios, where it surpasses all previous state-of-the-art methods including Learning-to-Cache~\cite{ma2024learning}, DiP-GO~\cite{zhu2024dip}, DeepCache~\cite{ma2024deepcache} and Diff-Pruning~\cite{fang2023structural}. This study is the first to systematically integrate the learning speed variations from the pretraining phase into the design of a post-training acceleration method, thereby providing a new solution for training-free diffusion model acceleration. We summarize our contributions as follows:
\begin{itemize}
\item We introduce a pruning strategy that aligns training-free / post-training acceleration with the varying learning speeds of diffusion pretraining without actual pretraining.
\item We propose a novel approach to identify stage-specific sparse networks, applying aggressive pruning during slow-learning phases and conservative pruning during fast-learning phases.
\item Our extensive experiments demonstrate state-of-the-art generation and acceleration performance, particularly at high pruning ratios, outperforming existing training-free methods by significant margins.
\end{itemize}

\section{Related work} 
\label{sec:related_work}

\paragraph{Efficient Diffusion Models.} Diffusion models exhibit exceptional generative performance but face high computational costs due to their iterative denoising process \cite{song2019generative}. Inference acceleration methods mainly aim to either reduce the number of sampling steps or optimize computations per step. Step-reduction approaches such as DDIM \cite{song2020denoising}, DPM-Solver \cite{lu2022dpm}, and Consistency Models \cite{song2023consistency} reformulate the diffusion process, while knowledge distillation approaches \cite{salimans2022progressive, meng2022distillation} transfer multi-step denoising capabilities from larger teacher models to compact students. Meanwhile, per-step optimization strategies employ architectural compression methods, including structural pruning \cite{fang2023pruning, kim2023architectural, zhang2024effortless}, model distillation \cite{meng2023distillation, kim2024bk}, and early stopping \cite{lyu2022accelerating, tang2024adadiff}. Additional techniques, such as quantization \cite{he2023ptqd, shang2023post, li2023q, li2024svdquant} and feature caching \cite{ma2024learning, ma2024deepcache, zhu2024dip}, further reduce computation and memory usage. Finally, trajectory stitching \cite{pan2024tstitch} combines models of varying complexities during inference stages without degrading generation.

\noindent{\bf Noise Schedule Optimization.}  Noise schedules are critical hyperparameters that directly influence learning speeds throughout the diffusion process. Prior studies \cite{nichol2021improveddenoisingdiffusionprobabilistic, lin2024common, podellsdxl} have demonstrated that these schedules significantly affect the training outcomes of diffusion models. For example, Choi et al. \cite{choi2022perception} assign weights to emphasize important training steps based on noise schedules, enhancing training performance. However, existing works often overlook that noise schedules similarly impact the inference stage, essentially an unguided extension of the training process. Leveraging insights from pretrained model dynamics, we propose utilizing the step-wise importance indicated by the noise schedule to accelerate and enhance the sampling process.

\begin{figure*}[ht]
    \centering
    \begin{subfigure}{0.55\linewidth}
    \includegraphics[width=\linewidth]{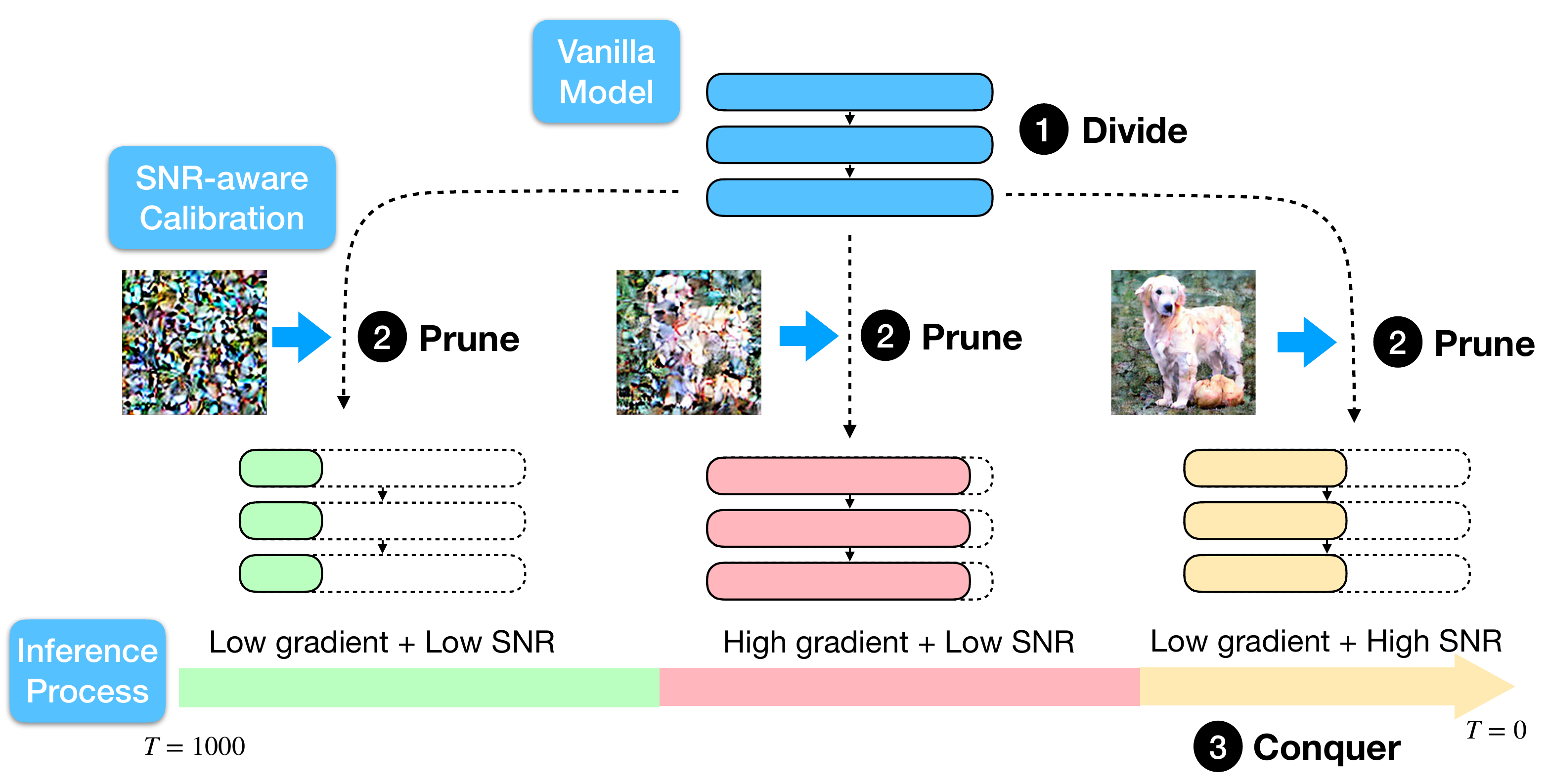}
    \caption{Main pipeline of Divide, Prune and Conquer.}
    \label{fig:DPC}
  \end{subfigure}
  \hfill
  \vrule width 0.8pt 
    \hspace{3pt}
  \begin{subfigure}{0.42\linewidth}
    \includegraphics[width=\linewidth]{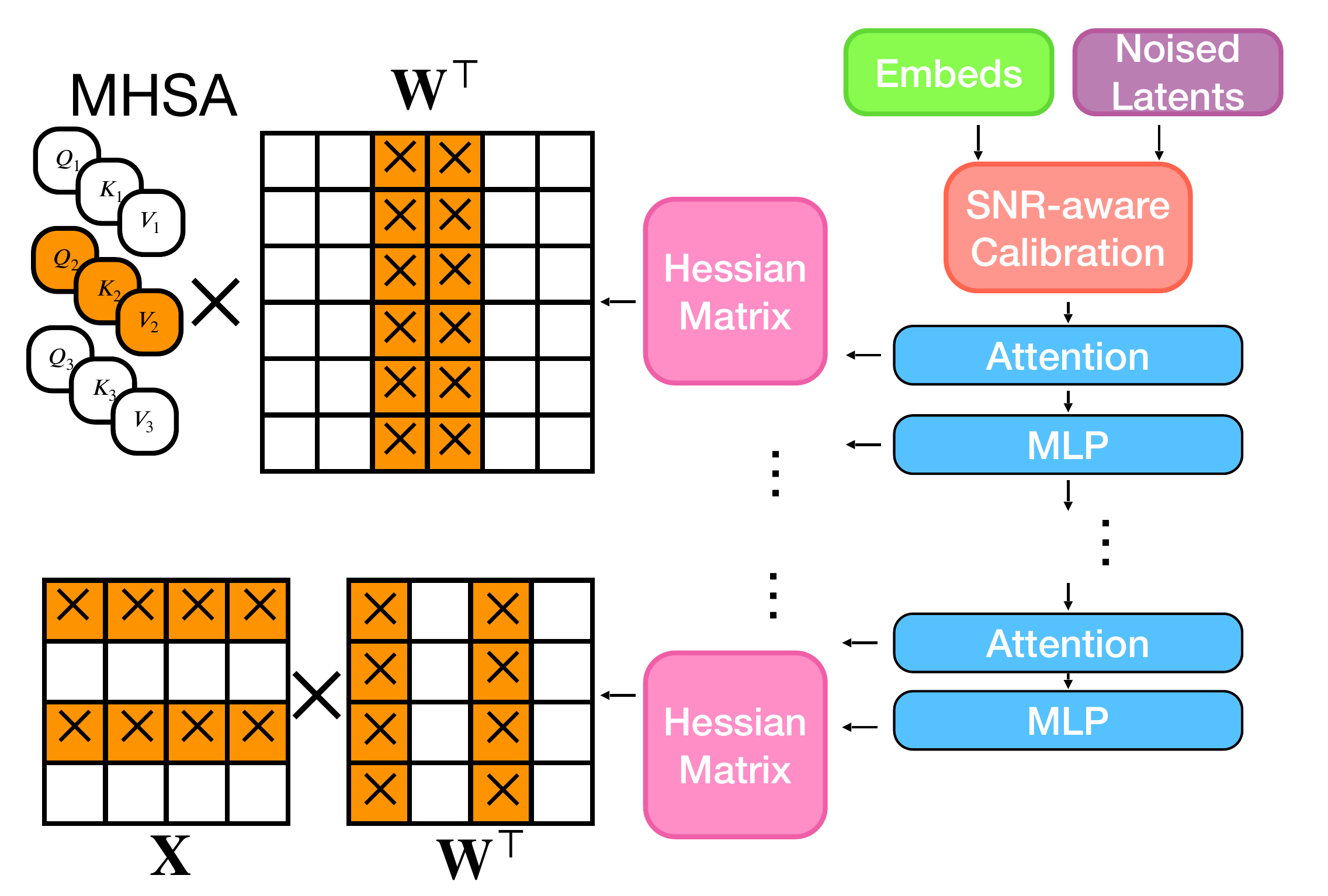}
    \caption{Second-order structural pruning for sub diffusion networks.}
    \label{fig:pruning}
  \end{subfigure}
  \vspace{-0.13in}
  \caption{\textbf{Overview of \em MosaicDiff.} (a) \textbf{Main framework:} We first divide the inference process into three distinct stages according to a quantitative analysis of pretraining dynamics. For each stage, we utilize SNR-aware calibration data to perform second-order structural pruning, obtaining subnetworks with varying degrees of sparsity. Finally, we integrate these subnetworks to enable efficient inference across all timesteps. (b) \textbf{Second-order structural pruning:} To practically implement pruning on diffusion models, we feed SNR-aware calibration data into the pretrained model, computing Hessian matrices for each Attention and MLP layer. We then derive saliency scores from these Hessians to prune less important weight columns, corresponding to heads in multi-head self-attention (MHSA) layers and neurons in intermediate MLP layers.}
  \vspace{-0.13in}
\end{figure*}

\noindent{\bf Structural Pruning.}  Structural pruning has been widely applied to large neural networks, such as large language models (LLMs) \cite{frantar2023sparsegpt, kurtic2023ziplm, tang2025darwinlm}, to efficiently accelerate inference by removing entire substructures (e.g., neurons, channels, or layers). However, pruning methods for diffusion models are still in their early stages due to their iterative nature and heightened sensitivity to parameter reduction \cite{peebles2023scalable}. Recent works, such as Diff-Pruning \cite{fang2023pruning}, utilize improved Taylor pruning to identify redundant structures, while EcoDiff \cite{zhang2024effortless} employs a training-based differentiable mask for model sparsification. However, these approaches still yield limited performance and are constrained to low sparsity levels, primarily due to their use of a uniformly pruned model across all diffusion timesteps. This approach overlooks the inherent step-wise importance dynamics in the diffusion sampling process, potentially missing further efficiency gains.
\section{\em MosaicDiff}
\label{sec:methodology}

\noindent{\bf Framework Overview.} 
Figure~\ref{fig:DPC} illustrates the main framework of {\bf{\em MosaicDiff}}, which contains three main phases {\em Divide, Prune and Conquer}.
Starting from a pre-trained vanilla large diffusion model, we need to determine both the pruning stages and the corresponding sparsity levels.
In the {\em Divide} phase, the inference trajectory is split into segments according to the strategy introduced in Section~\ref{Sec:Stage}. Within each segment, a {\em Prune} step applies second-order structural pruning guided by SNR-aware calibration data (Sections~\ref{Sec:Calib_Data} and~\ref{Sec:Pruning}). Finally, the {\em Conquer} step merges these pruned sub-networks for the final sampling, ensuring that the accelerated inference remains consistent with the original model's trajectory.

\subsection{Preliminary}
Diffusion models comprise a forward training process and a reverse sampling process. The forward process learns the features in images by gradually adding Gaussian noise to them on the basis of a Markov chain. Generally, diffusion models add noise according to a pre-defined and fixed hyper-parameter schedule $\beta_1,...,\beta_T.$ Thus, we can formulate the forward process as:
\begin{equation}
    q(x_{1:T} \mid x_0) := \prod_{t=1}^T q(x_t \mid x_{t-1}),
    \label{Markov forward process}
\end{equation}
\begin{equation}
    q(x_t \mid x_{t-1}) := \mathcal{N} \Bigl(x_t;\,\sqrt{1-\beta_t}x_{t-1},\,\beta_t\mathbf{I} \Bigr).
    \label{normal forward process}
\end{equation}
If further define $\alpha_t := 1 - \beta_t$ and $\bar{\alpha}_t := \prod_{s=1}^t\alpha_s$, we can sample $x_t$ at an arbitrary timestep $t$ using a closed form:
\begin{equation}
    q(x_t \mid x_0) = \mathcal{N} \Bigl(x_t;\,\sqrt{\bar{\alpha}_t}x_0,\,(1-\bar{\alpha}_t)\mathbf{I} \Bigr),
    \label{closed form normal forward process}
\end{equation}
\noindent
which equals to:
\begin{equation}
    x_t(x_0,\,\epsilon) = \sqrt{\bar{\alpha}_t}x_0 + \sqrt{1-\bar{\alpha}_t}\epsilon, \quad 
\, \epsilon \sim \mathcal{N}(\mathbf{0},\mathbf{I}).
    \label{closed form forward process}
\end{equation}
Thus, we can calculate signal-to-noise ratio (SNR) of the image at the timestep $t$ by:
\begin{equation}
    \text{SNR}(t) = \frac{\bar{\alpha}_t}{1-\bar{\alpha}_t}.
    \label{signal-to-noise ratio}
\end{equation}

The reverse process of diffusion models generates images by gradually denoising pure Gaussian noises from a distribution defined as $p(x_T)= \mathcal{N}(x_T;\,\mathbf{0},\mathbf{I})$ according to a similar Markov chain:
\begin{equation}
    p_\theta(x_{0:T}) := p(x_T) \prod_{t=1}^Tp_\theta(x_{t-1} \mid x_t),
    \label{Markov reverse process}
\end{equation}
\begin{equation}
    p_\theta(x_{t-1} \mid x_t) := \mathcal{N} \Bigl(x_{t-1};\,\mu_\theta(x_t,t),\,\Sigma_\theta(x_t,t)\Bigr).
    \label{normal reverse process}
\end{equation}
Set $\Sigma_\theta(x_t,t) = \sigma^2\mathbf{I}$, where $\sigma^2$ can be directly calculated through $\beta_t$, and reparameterize $\mu_\theta(x_t,t)$ as:
\begin{equation}
    \mu_\theta(x_t,t) = \frac{1}{\sqrt{\alpha_t}}\Bigl(x_t - \frac{\beta_t}{\sqrt{1-\bar{\alpha}_t}}\epsilon_\theta(x_t,t)\Bigr).
    \label{reparameterize mu}
\end{equation}
Then what neural networks really predict at timestep $t$ is the noise $\epsilon_\theta(x_t,t)$. The loss function is:
\begin{equation}
    \mathcal{L}(\theta)= \mathbb{E}_{x_0,\epsilon}\Bigl[\|\epsilon - \epsilon_\theta(\sqrt{\bar{\alpha}_t}x_0 + \sqrt{1-\bar{\alpha}_t}\epsilon,t)\|^2 \Bigr].
    \label{diffusion models' loss function}
\end{equation}

\subsection{Stage Division for Reverse Process}
\label{Sec:Stage}

\begin{figure}[t]
    \centering
    \includegraphics[width=\linewidth]{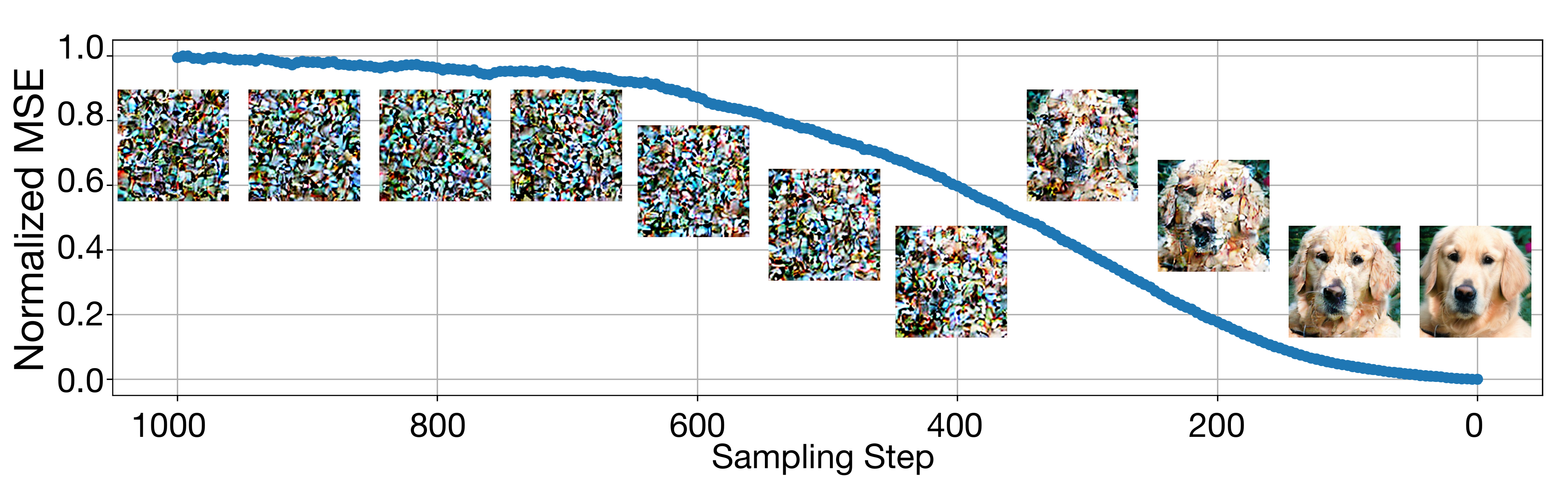}
    \vspace{-0.2in}
    \caption{Change in image MSE over sampling steps: In the early stage $T \in (600,1000)$, the MSE decreases slowly with images remaining largely noisy, in the middle stage $T \in (200,600)$, denoising accelerates and images converge rapidly, and in the final stage $T \in (0,200)$, MSE reduction slows, indicating only subtle perceptual refinements.}
    \label{fig:image MSE change}
    \vspace{-0.15in}
\end{figure}

\noindent\textbf{Empirical Observation.} A key observation driving {\bf \em MosaicDiff} is that diffusion pretraining exhibits distinct phases of learning speed, which are reflected in the reverse denoising steps. Prior research has empirically shown that the importance of different timesteps in diffusion models varies significantly. Some studies \cite{choi2022perception, pan2024tstitch, nichol2021improveddenoisingdiffusionprobabilistic} emphasize this by analyzing the evolution of the SNR throughout the process, while others reach a similar conclusion by comparing feature similarities along the generation trajectory. 

To harness these differences, we partition the reverse process into multiple stages based on timestep intervals, assigning each stage its own tailored sparsity budget. This strategic segmentation allows us to align the pruning intensity with the learning dynamics observed during pretraining. Our approach specifically focuses on monitoring the changes in Mean Squared Error (MSE) between the intermediate latents representations $\hat{x}_t$ and the final output latents $\hat{x}_0$, i.e.:
\begin{equation}
    \text{MSE}(t) = \frac{1}{d}\|\hat{x}_t - \hat{x}_0\|_2^2
    \label{generation MSE},
\end{equation}
where $d$ is the dimension of latents $\hat{x}_t$ and $\hat{x}_0$.
Using outputs from the DiT-XL/2 model as an example, we plot the MSE trend across sampling steps $t$. As shown in Figure \ref{fig:image MSE change}, the MSE decreases gradually during the early stages of generation, with the associated gradients remaining nearly constant. In contrast, during the intermediate stages, the MSE rapidly diminishes while the gradients increase significantly. Finally, in the later steps of the reverse process, the rate of MSE reduction slows again, and the gradients subside to lower levels. We claim that the expected MSE and gradients for each step can be derived via a closed-form solution, which provides theoretical support as the following theorem for our stage-specific pruning strategy.
\begin{theorem}
With the $\bar{\alpha}_t$ from the noise scheduler, the expectation of MSE and gradient can be formulated as :
\begin{align}
\mathbb{E}\Bigl[\text{MSE}(t)\Bigr] &= \frac{1}{d}\Bigl[(1-\sqrt{\bar{\alpha}_t})^2\|\hat{x}_0\|_2^2 + (1 - \bar{\alpha}_t)\|\mathbf{I}\|_2^2\Bigr], \\
    \mathbb{E}\Bigl[\text{Grad}(t)\Bigr] \!&=\! \frac{1}{d}\Bigl[(\delta_t + 2(\sqrt{\bar{\alpha}_{t-1}}-\sqrt{\bar{\alpha}_t}))\|\hat{x}_0\|_2^2 - \delta_t\|\mathbf{I}\|_2^2\Bigr].
\end{align}
\end{theorem}
Define $\delta_t := \bar{\alpha}_t - \bar{\alpha}_{t-1}$. Detailed derivations and proofs are provided in Appendix. In Figure~\ref{fig:MSE comparison}, we compare the closed-form MSE and gradient curves with their empirical counterparts. The close alignment between the theoretical and observed curves validates our equations and supports our approach.
\begin{figure}[t]
    \centering
    \vspace{-0.15in}
    \includegraphics[width=1\linewidth]{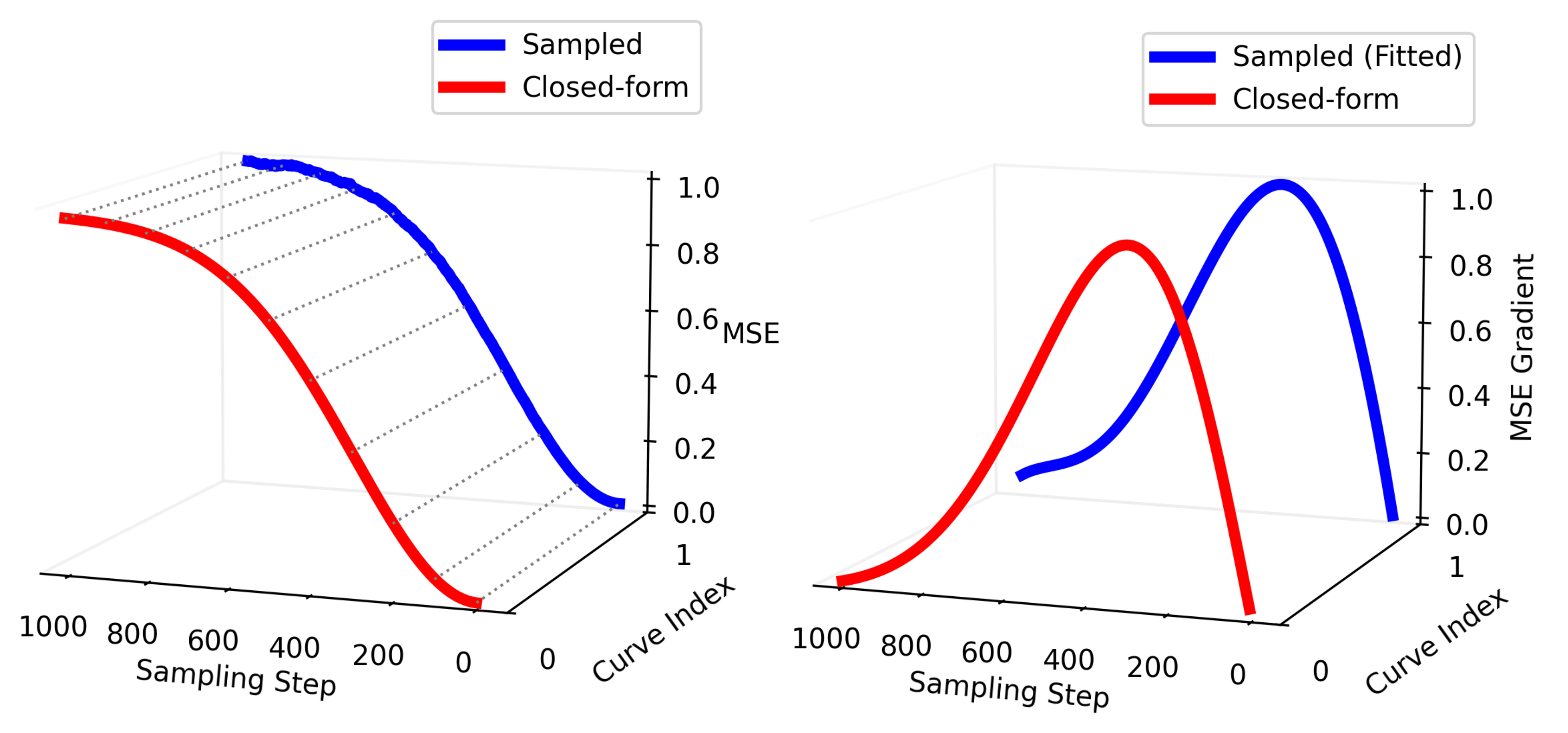}
    \vspace{-0.3in}
    \caption{MSE and gradient curves comparison under Linear Schedule. \textit{Left}: MSE calculated from our closed-form approximation closely matches the sampled results. \textit{Right}: Gradients derived from our closed-form expression align with empirically sampled gradients. }
    \label{fig:MSE comparison}
    \vspace{-0.1in}
\end{figure}

\begin{figure}[t]
    \centering
    \begin{subfigure}{0.48\linewidth}
    \includegraphics[width=\linewidth]{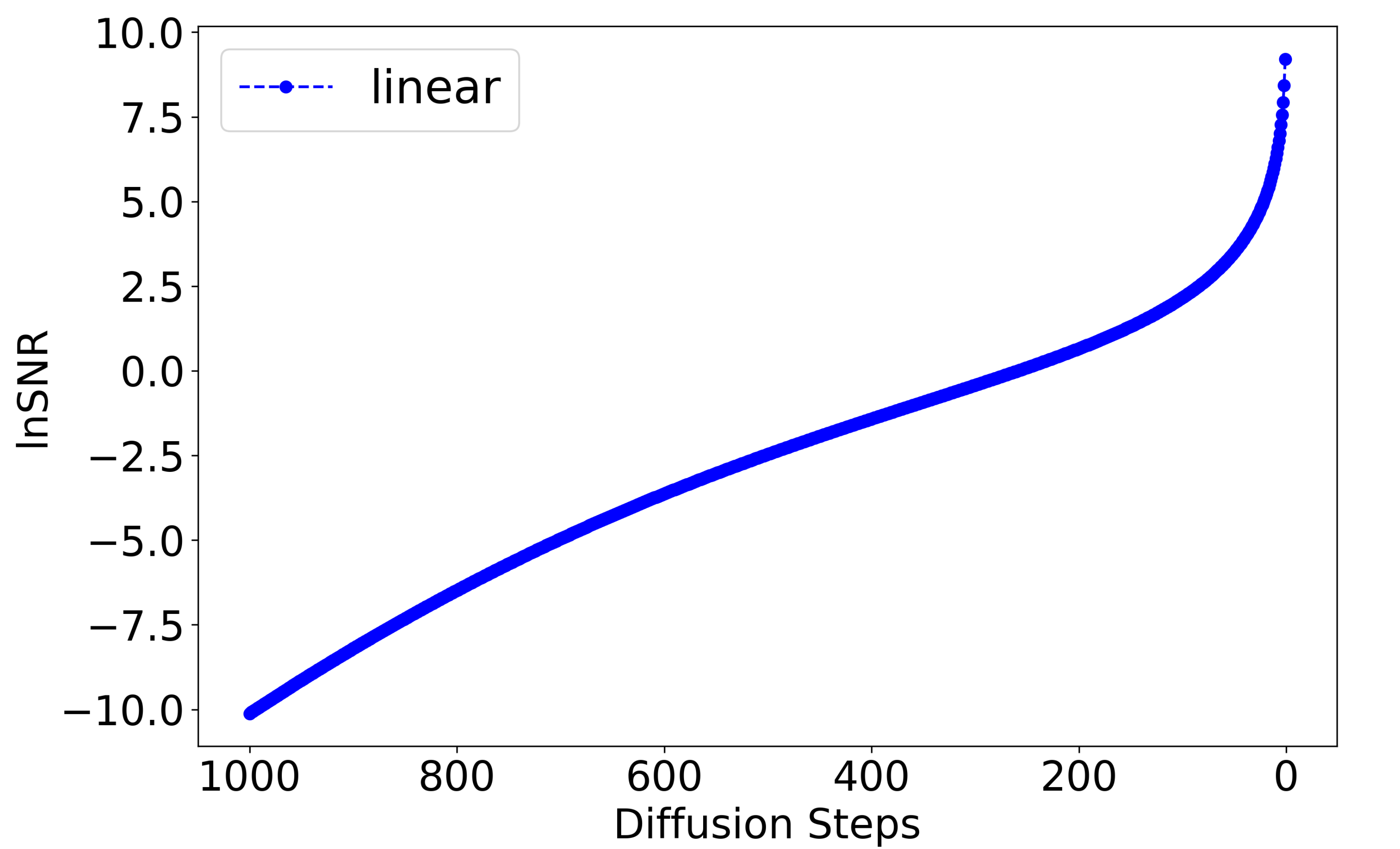}
    \caption{SNR trend of linear schedule.}
    \label{fig:SNR change}
  \end{subfigure}
  \hfill
  \begin{subfigure}{0.48\linewidth}
    \includegraphics[width=\linewidth]{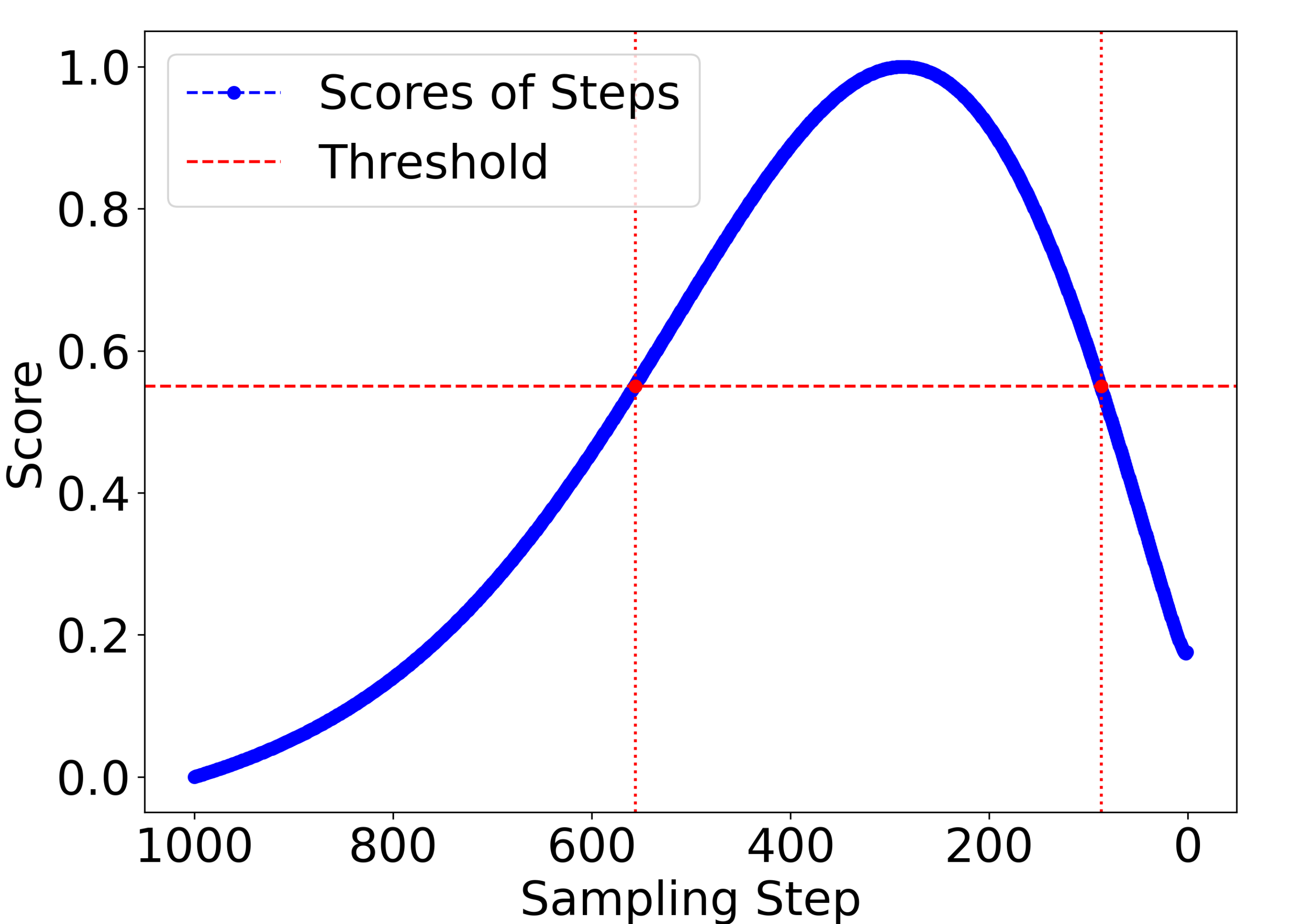}
    \caption{Final scores of sampling steps.}
    \label{fig:Final scores}
  \end{subfigure}
  \caption{Influence of SNR on Final Scores. (a) Change in SNR across sampling steps, showing a sharp increase during the final steps. (b) Final scores computed combining SNR. A threshold of $M=0.55$ clearly divides the curve into three stages.}
  \vspace{-1em}
\end{figure}

\noindent\textbf{Quantitative Importance Scores.} We define an importance score, $score(t)$, for each sampling step to quantify its contribution during inference. Our formulation is based on the above observation that the gradient of MSE is indicative of convergence speed - the larger the gradient, the more rapidly the step approaches the final output, ensuring a lower sparsity level $s$ in our pruning algorithm. Moreover, as shown in Figure \ref{fig:SNR change}, the SNR significantly increases during the final sampling steps. Although the gradients in these stages are comparably small to those in the early steps, even slight changes in the MSE can have a disproportionately large impact on the final image quality. To capture this effect, we incorporate the SNR (from Equation (\ref{signal-to-noise ratio})) into our importance score as follows:
\begin{equation}
    score(t) = \mathbb{E}\Bigl[\text{Grad}(t)\Bigr] + \lambda \ln \text{SNR}(t)
    \label{final score}
\end{equation}
where $\lambda$ is a hyperparameter that controls the influence of the SNR, making the final $score(t)$ entirely dependent on the noise schedule $\bar{\alpha}_t$.
Figure \ref{fig:Final scores} shows the final score curves for DiT-XL/2 under a linear schedule. The results of SDXL under a scaled linear schedule are in Appendix. 

\noindent \textbf{Stage and Sparsity Decider.} After computing the final $score$ of a diffusion model based on $\bar{\alpha}_t$, we define:
\begin{equation}
    threshold := M \cdot \max_t(score(t)),
    \label{threshold}
\end{equation}
where $M \in (0,1)$ is a hyper-parameter. As illustrated in Figure \ref{fig:Final scores}, this threshold divides the score curve into three segments, each corresponding to a pruning stage with a fixed sparsity $s_i$, where $i \in \{0,1,2\}$. We then compute the average score $\overline{score}_i$ for each segment and set:
\begin{equation}\label{sparsity}
    s_i \propto 1 - \overline{score}_i.
\end{equation}
Intuitively, segments with higher average scores receive lower sparsity, preserving more parameters in timesteps deemed critical. Based on Equation (\ref{sparsity}), we assign each specific sparsity level to each segment by leveraging both our experimental results and the observed learning dynamics presented above. Details are presented in the Appendix.

\subsection{SNR-Aware Calibration Dataset}
\label{Sec:Calib_Data}
After partitioning the reverse process into stages and determining the corresponding sparsities, we build an SNR-aware calibration dataset for pruning. Our objective is to have each model specialize in a specific stage, excelling at inferring on latents that fall within a targeted SNR range. To achieve this, we first resize each image from a standard calibration set (e.g., ImageNet-1K) and encode it into a latent representation.
Next, we randomly select a timestep $t$ within the stage of interest and add noise to the latent according to Equation~(\ref{closed form forward process}) to ensure that the resulting latent meets the desired SNR value, i.e., $\text{SNR}(t)$. The noised latent, along with its associated timestep $t$ and the corresponding class label or caption, is then packaged to be considered in the Hessian computation for pruning.

For models using classifier-free guidance (CFG) during inference, we mimic unconditional generation by providing a null-label calibration example alongside each noised latent. Specifically, when CFG is enabled, we duplicate the latent and $t$ (concatenating them) and pair this with a null label to ensure the calibration process accurately reflects the inference setup.
This comprehensive calibration approach, resizing, encoding, random timestep selection, noise addition, and appropriate pairing with labels (or null labels under CFG) ensures that the model receives precise Hessian information for each linear layer. This, in turn, guides the following stage-specific second-order structural pruning for optimal acceleration.

\subsection{Second-Order Structural Pruning}
\label{Sec:Pruning}
In contrast to existing compression methods, we employ the Hessian matrix to assess the importance of substructures more precisely for different sampling stages, resulting in a training-free, fine-tuning-free pruning algorithm that can be seamlessly applied to our accelerating diffusion framework. 

\noindent\textbf{Problem Formulation.} Given a linear layer with calibration input $\mathbf{X} \in \mathbb{R}^{b \times n}$ and weight $\mathbf{W} \in \mathbb{R}^{m \times n}$, our method aims to find the compressed weight $\widehat{\mathbf{W}}$ at the pruning sparsity $s$, which causes the least error compare with the original output, evaluated by the square error:
\begin{equation}
    \text{argmin}_{\widehat{\mathbf{W}}}\|\mathbf{X}\widehat{\mathbf{W}}^\top-\mathbf{X}\mathbf{W}^\top\|_2^2
    \label{pruning error}
\end{equation}
This objective can be decomposed into independent minimization jobs across the $m$ rows of $\mathbf{W}$. However, since we focus on structural pruning, the indices we remove at all rows should be the same, i.e.\ we should prune entire columns. Define a column mask $\mathbf{M} \in \mathbb{R}^n$, which only contains value 0 and 1. $\mathbf{M}_i = 1$ means column $i$ need to be pruned and $\mathbf{M}_i = 0$ means column $i$ to be preserved, where $i \in [0,n).$ We need to prune weight $\mathbf{W}$ to sparsity $s$, so $\|\mathbf{M}\|_1 = \lfloor s\cdot n \rfloor.$ Therefore, our goal becomes to find the optimal mask $\mathbf{M}$, so that pruning out $\mathbf{W}_{:,\mathbf{M}}$ will result in the least error.

\noindent\textbf{Saliency Score and Weight Update.} After second-order derivative, the Hessian matrix for the $\ell_2$-minimization problem in Equation (\ref{pruning error}) can be calculated as $\mathbf{H}=\mathbf{X}^\top\mathbf{X} \in \mathbb{R}^{n \times n}$. We further solve Equation (\ref{pruning error}) by extending formulas from Optimal Brain Surgeon (OBS) \cite{298572}. Then, we can obtain the saliency score of the optimal mask:
\begin{equation}
    \text{argmin}_{\mathbf{M}} \sum_{i=0}^{m-1} \mathbf{W}_{i,\mathbf{M}} \cdot \Bigl({\mathbf{H}_{\mathbf{M},\mathbf{M}}^{-1}}\Bigr)^{-1} \cdot \mathbf{W}_{i,\mathbf{M}}^\top
    \label{saliency score}
\end{equation}
After eliminating the columns according to the mask, we can also compute an update $\delta$ for the remaining weights to compensate for pruning to further reduce errors:
\begin{equation}
    \delta = - \mathbf{W}_{:,\mathbf{M}} \cdot \Bigl({\mathbf{H}_{\mathbf{M},\mathbf{M}}^{-1}}\Bigr)^{-1} \cdot \mathbf{H}_{\mathbf{M},:}^{-1}
    \label{compensation}
\end{equation}
Lastly, we multiply the weight matrix with the mask $\mathbf{W} \odot ( \mathbf{1-M})$ to ensure the pruned columns are exactly zeros.

\noindent\textbf{Pruned Structures.} In our work with diffusion models, both the Transformer and U-Net architectures are examined under two pruned configurations: (1) head pruning within the multi-head self-attention, and (2) the reduction of the intermediate dimension in MLP modules.

\begin{table*}[t]
  \centering
  \caption{Result comparison of {\bf{\em MosaicDiff}} and other baselines on DiT-XL/2 using DDIM sampler. {\bf{\em MosaicDiff}} consistently outperforms all baselines across different sampling step configurations in terms of FID, IS, and Precision, while maintaining competitive Recall. Moreover, {\bf{\em MosaicDiff}} achieves the highest speedup and the lowest MACs, demonstrating its superior efficiency in accelerated sampling.}
\vspace{-1em}
  \label{tab:main_dit_cmp_sota}
  \resizebox{0.9\textwidth}{!}{
  \begin{tabular}{l|c|ccc|cccc}
    \toprule
    \textbf{Method} & 
    \textbf{Steps} & 
    \textbf{MACs (T)} & 
    \textbf{Latency (s)} & 
    \textbf{Speedup} & 
    \textbf{IS} \(\uparrow\) &   
    \textbf{FID} \(\downarrow\) & 
    \textbf{Precision} \(\uparrow\) & 
    \textbf{Recall} \(\uparrow\)\\
    \midrule
    Vanilla DiT-XL/2 & 250 & 28.61 & 19.20 & 1.00\(\times\) & 243.4 & 2.14 & 80.70 & 60.57 \\
    \midrule
    Vanilla DiT-XL/2 & 50  & 5.72  & 3.83  & 1.00\(\times\) & 238.6 & 2.26 &80.16 & \textbf{59.89} \\
    Diff-Pruning-0.3 & 50  & 4.10  & 2.98  & 1.29\(\times\) & 4.68 & 180.76 &7.24 & 20.26 \\
    Learning-to-Cache & 50  & 4.36  & 3.01  & 1.27\(\times\) & 244.1 & 2.27 & 80.94 & 58.76 \\
    \rowcolor{cyan!10}
    {\em MosaicDiff} (Ours) Pruned-0.33 & 50  & 3.92  & 2.90  & 1.32\(\times\) & \textbf{267.8} & \textbf{2.24} & \textbf{82.01} & 57.31 \\
    \midrule
    Vanilla DiT-XL/2 & 20  & 2.29  & 1.53  & 1.00\(\times\) & 223.5 & 3.48 & 78.76 & \textbf{57.07} \\
    Diff-Pruning-0.3 & 20  & 1.64 & 1.20  & 1.28\(\times\) & 2.99 & 223.80 & 2.92 & 12.38 \\
    Learning-to-Cache& 20  & 1.78  & 1.23  & 1.24\(\times\) & 227.0 & 3.46 & 79.15 & 55.62 \\
    \rowcolor{cyan!10}
    {\em MosaicDiff} (Ours) Pruned-0.30 & 20  & 1.64  & 1.20  & 1.28\(\times\) & \textbf{266.7} & \textbf{3.20} &\textbf{81.13} & 53.67 \\
    \midrule
    Vanilla DiT-XL/2 & 10  & 1.14  & 0.77  & 1.00\(\times\) & 158.3 & 12.38 & 66.78 & \textbf{52.82} \\
    Diff-Pruning-0.3 & 10  & 0.82  & 0.63  & 1.22\(\times\) & 2.14 & 270.26 & 0.93 & 9.53 \\
    Learning-to-Cache & 10  & 1.04  & 0.69  & 1.12\(\times\) & 156.3 & 12.79 & 66.21 & 52.15 \\
    \rowcolor{cyan!10}
    {\em MosaicDiff} (Ours) Pruned-0.30 & 10  & 0.79  & 0.58  & 1.33\(\times\) & \textbf{174.6} & \textbf{12.28} & \textbf{66.95} & 49.40 \\
    \bottomrule
  \end{tabular}
  }
  \vspace{-1em}
\end{table*}
\begin{table}[t]
  \centering
  \caption{Result comparison with DiP-GO on DiT-XL/2. }
  \label{tab:main_dit_cmp_sota_large_step}
  \vspace{-1em}
  \resizebox{0.4\textwidth}{!}{
  \begin{tabular}{l|c|c|c}
    \toprule
    \textbf{Method} & 
    \textbf{Steps} & 
    \textbf{MACs (T)} & 
    \textbf{FID} \(\downarrow\) \\
    \midrule
    Vanilla DiT-XL/2   & 100 & 11.86 & 3.17 \\
    DiP-GO Pruned-0.6  & - & 11.86 & 3.01\\
    \rowcolor{cyan!10}
    {\em MosaicDiff} (Ours) Pruned-0.25  & 100 & 8.52 & \textbf{2.67}\\ 
    \midrule
    Vanilla DiT-XL/2   & 70 & 8.30 &  3.35\\
    DiP-GO Pruned-0.75  & - & 6.77 &  3.14\\
    \rowcolor{cyan!10}
    {\em MosaicDiff} (Ours) Pruned-0.25  & 70 & 5.99 &  \textbf{3.01}\\ 
    \bottomrule
  \end{tabular}}
  \vspace{-1em}
\end{table}

\section{Experiments}
\subsection{Setup} 

\noindent \textbf{Models.} We validate our methods on two mainstream latent diffusion models, leveraging both transformer \cite{vaswani2017attention} and U-Net \cite{ronneberger2015u} architectures: 1) DiT \cite{peebles2023scalable} is a widely adopted transformer-based diffusion model available in multiple parameter configurations. We focus on pruning the 256 $\times$ 256 linear scheduled DiT-XL/2 variant, which consists of 675 million parameters, to demonstrate the effectiveness of our approach. 2) SDXL-base-1.0 \cite{podellsdxl} is a state-of-the-art U-Net-based text-to-image diffusion model with 2.6 billion parameters. We employ this model to present the scalability of our methods on large-scale architectures and prove the compatibility with scaled-linear scheduled models.

\noindent \textbf{Datasets and Metrics.} For DiT, following previous work, we conduct experiments on ImageNet-1K \cite{deng2009imagenet} at the resolution of 256 $\times$ 256. For SDXL, we use the MS COCO 2017 \cite{lin2014microsoft} for quantitative evaluation. We follow the evaluation of DiT, we generate 50,000 images and evaluate their quality using the Fréchet Inception Distance (FID) \cite{heusel2017gans}, computed with ADM’s TensorFlow evaluation suite \cite{dhariwal2021diffusion}, consistent with prior work. Additionally, we report Inception Score (IS) and Precision-Recall as complementary metrics. For SDXL, we include FID-5K calculated by torchmetrics to assess image quality, CLIP-Score \cite{radford2021learning} calculated by ViT-B-16 to measure text-image alignment, and SSIM \cite{1284395} to quantify output differences compared to the original model outputs. The MACs is evaluated using PyTorch-OpCounter, and the latency is tested when generating 8 images on a single RTX 4090, which we conduct five times and compute the average.

\noindent \textbf{Baselines.} We compare our method against state-of-the-art sampling acceleration techniques. For DiT, we first compare our method with Diff-Pruning \cite{fang2023structural} by utilizing their official implementation. We also compare against DiP-GO \cite{zhu2024dip}, a training-based compression approach, to highlight our effectiveness at larger sampling steps when using the conventional DDPM \cite{ho2020denoising}. We report the results of DiP-GO from their original paper. Moreover, to show the superior performance of our method on acceleration in a training-free manner, we compare our approach with Learning-to-Cache (L2C) \cite{ma2024learning}, a caching-based method that requires training. We also report the results of L2C from their paper.
For SDXL, we first compare with pruning approaches Diff-Pruning and Eco-Pruning \cite{zhang2024effortless}.  Because Eco-Pruning targets memory reduction more than compute speed, we briefly include its reported results from the original paper. Moreover, in addition to pruning methods, we also compare with the training-free caching approach DeepCache \cite{ma2024deepcache}.

\noindent \textbf{Implementation Details.} On DiT, we randomly choose 1024 images from ImageNet-1K as calibration for each pruning stage. On SDXL, we randomly select 1024 images and paired captions from MS COCO 2017 training dataset as calibration. For both models, we select threshold $M = 0.55$ and influence of SNR $\lambda = 0.01$. All experiments are conducted on NVIDIA RTX 4090 GPU.

\subsection{Main results}

\noindent \textbf{Result Comparison on DiT.} 
We present a comprehensive comparison between our proposed method and baseline approaches, as summarized in Table \ref{tab:main_dit_cmp_sota} and Table \ref{tab:main_dit_cmp_sota_large_step}. First, we compare our method with Diff-Pruning under identical sparsity constraints. Without additional fine-tuning, our method significantly outperforms Diff-Pruning, achieving an FID of 2.24 compared to 180.76 obtained by Diff-Pruning using 50 sampling steps. As the number of sampling steps decreases, the performance gap widens further. Specifically, with only 20 and 10 sampling steps, Diff-Pruning attains FID scores of 223.8 and 270.26, respectively, while our {\bf{\em MosaicDiff}} achieves substantially lower FID scores of 3.20 and 12.28.

Moreover, our method surpasses both the vanilla DiT and DiP-GO at 100 and 70 sampling steps, achieving lower MACs and FID scores, thereby producing higher-quality images with reduced computational cost. Furthermore, {\bf{\em MosaicDiff}} consistently outperforms L2C across all evaluated step counts, achieving superior results in IS, FID, and Precision. Notably, under 20-step sampling using the fast DDIM sampler, our method achieves an FID of 3.20, outperforming both L2C and the uncompressed baseline model. While the original DiT achieves slightly higher Recall, the minor reductions observed in both L2C and {\bf{\em MosaicDiff}} indicate a negligible impact on generative diversity. These results highlight the superior performance of {\bf{\em MosaicDiff}} and demonstrate the effectiveness of the proposed approach.
\begin{table}[t]
  \centering
  \caption{ Comparison of {\bf{\em MosaicDiff}} with existing pruning methods on SDXL at different sparsity levels.}
  \vspace{-1em}
  \label{tab:main_sxdl_cmp_pruning}

  \resizebox{0.4\textwidth}{!}{
\begin{tabular}{l|c|ccc}
\toprule
\textbf{Method} &
 \textbf{Sparsity} &
    \textbf{FID}\(\downarrow\)         & \textbf{CLIP} \(\uparrow\)       & \textbf{SSIM}  \(\uparrow\)      \\ \midrule
Vanilla SDXL     & 0     & 24.90       & 0.32              & 1                    \\ \midrule
Diff-Pruning & 10\%  & 108.96            & 0.22                 & 0.31                 \\
EcoDiff      & 10\%  & 33.75             & 0.31                 & 0.53                 \\
\rowcolor{cyan!10}
{\em MosaicDiff} (Ours)         & 10\%  & \textbf{23.18}             & \textbf{0.32}                & \textbf{0.67}               \\ \midrule
Diff-Pruning & 20\%  & 404.87            & 0.05                 & 0.26                 \\
EcoDiff      & 20\%  & 34.41             & 0.31                 & 0.5                  \\
\rowcolor{cyan!10}
{\em MosaicDiff} (Ours)         & 20\%  & \textbf{23.79}             & \textbf{0.32}              & \textbf{0.64}              \\ \midrule
EcoDiff      & 24\%  &  61.00             & -                &  -                \\
\rowcolor{cyan!10}
{\em MosaicDiff} (Ours)        & 30\%   &  \textbf{28.37}             &   0.31               &  0.53          \\ \bottomrule

\end{tabular}
}
\vspace{-1em}
\end{table}
\begin{table}[t]
  \centering
  \caption{Comparison of {\bf{\em MosaicDiff}} Pruned-0.15 with DeepCache on SDXL across different sampling step configurations.} 
  \vspace{-1em}
  \label{tab:main_sdxl_cmp_deepcache}
  \resizebox{0.35\textwidth}{!}{
  \begin{tabular}{l|c|c|c}
    \toprule
    \textbf{Method} & 
    \textbf{Steps} & 
    \textbf{MACs (T)} & 
    \textbf{FID} \(\downarrow\) \\
    \midrule
    Vanilla SDXL   & 50 & 159.60 & 24.90 \\
    {\em MosaicDiff} (Ours)  & 50 & 135.66 & \textbf{23.73}\\
    \midrule
    DeepCache-N=2  & 50 & 93.23 & 24.88\\
    {\em MosaicDiff} (Ours)  & 25 & 72.04 & \textbf{24.04}\\
    \midrule
    DeepCache-N=3  & 50 & 59.66 & 24.67\\
    {\em MosaicDiff} (Ours)  & 20 & 57.63 & \textbf{24.17}\\ 
    \midrule
    DeepCache-N=5  & 50 & 37.26 & 24.43\\
    {\em MosaicDiff} (Ours)  & 10 & 28.82 & \textbf{24.32}\\
    \bottomrule
  \end{tabular}
  }
  \vspace{-1.5em}   
\end{table}

\noindent \textbf{Result Comparison on SDXL.} To show the scalability and compatibility
on other architectures, we evaluate our method on a larger U-Net based model, SDXL and provide the result comparison in Table \ref{tab:main_sxdl_cmp_pruning} and \ref{tab:main_sdxl_cmp_deepcache}. Without further fine-tuning, Diff-Pruning continues to underperform across all metrics, achieving FID scores of 108.96 and 404.87 at sparsity of 10\% and 20\%, respectively. In contrast, our training-free method consistently outperforms EcoDiff in terms of FID, CLIP, and SSIM. These results indicate that our method not only produces higher-quality images but also maintains better semantic alignment with textual prompts and better preserves the visual similarity to the original SDXL baseline.
We further compare our approach with DeepCache in Table~\ref{tab:main_sdxl_cmp_deepcache}. To ensure a fair comparison with DeepCache’s aggressive step-caching strategy, we evaluate our method at reduced sampling steps. As shown, our approach achieves lower FID scores at all step configurations while also requiring fewer MACs. These results highlight the superior efficiency and improved image quality delivered by our method on the SDXL architecture.

\subsection{Ablation study}
\noindent \textbf{SNR-Aware Calibration Dataset.} In this section, we highlight the importance of incorporating our SNR-aware calibration dataset. As shown in Table~\ref{tab:calibration}, utilizing SNR-aware calibration leads to a substantial improvement in generation quality, boosting the IS by over 40 points and reducing the FID by 0.74. Additionally, both Precision and Recall benefit from the SNR-aware calibration, demonstrating its effectiveness in enhancing both fidelity and diversity. These results clearly underscore the importance of leveraging SNR-aware data during the calibration process to achieve superior overall performance.
\begin{table}[t]
  \centering
  \caption{Ablation study on SNR-aware calibration dataset.}
  \vspace{-1em}
  \label{tab:calibration}
  \resizebox{0.4\textwidth}{!}{
  \begin{tabular}{l|cccc}
    \toprule
    \textbf{Calibration Dataset} & 
    \textbf{IS} \(\uparrow\)& 
    \textbf{FID} \(\downarrow\)&
    \textbf{Precision} \(\uparrow\)&
    \textbf{Recall}  \(\uparrow\)\\
    \midrule
    w/o SNR-aware &227.3	&3.96	&76.42	&53.73  \\
    w/ SNR-aware  &\textbf{266.1}	&\textbf{3.22} &	\textbf{81.48} &	\textbf{57.18} \\
    \bottomrule
  \end{tabular}}
  \vspace{-1em}
\end{table}
\begin{table}[t]
  \centering
  \label{tab:alation stage division}
  \caption{Ablation study on the stage division strategy.}
  \vspace{-1em}
  \label{tab:stage}
  \resizebox{0.4\textwidth}{!}{
  \begin{tabular}{l|cc|cc}
    \toprule
    \textbf{Strategy} & 
    \textbf{Divider 1} & 
    \textbf{Divider 2}&
    \textbf{IS}\(\uparrow\) &
    \textbf{FID} \(\downarrow\)\\
    \midrule
    None  &- &	-&	68.5&	40.89 \\
    Uniform  & 667 &	333&	80.6&	33.21  \\
    $M = 0.45$ & 600 & 50 & 253.2 & 3.36 \\
    \rowcolor{cyan!10}
    $M = 0.55$ & 550 & 100 & \textbf{266.7} & \textbf{3.20} \\
    $M = 0.70$ & 500 & 130 & 242.6 & 3.71 \\
    
    \bottomrule
  \end{tabular}}
  \vspace{-1.5em}
\end{table}

\noindent \textbf{Stage Division and Choice of $M$.} We emphasize the importance of dividing the denoising process into distinct stages based on the model’s sensitivity at different timesteps. We compare the proposed stage division (Ours) with two alternatives: no stage division (None) and uniform stage division and choice of $M$ in Equation \ref{threshold}. As shown in Table~\ref{tab:stage}, applying uniform sparsity across all timesteps without any stage division leads to a significantly degraded FID of 40.89. Performance improves when introducing a simple uniform stage division, but our proposed stage division achieves the best results across all metrics. Notably, in all cases, the overall sparsity is kept constant at 0.3 and sampling steps are 20 on DiT-XL/2, highlighting the importance of aligning the pruning strategy with the noise schedule to maximize performance.

\noindent \textbf{Sparsity Selection.} In this part, we compare the performance of different sparsity choices at every stage according to Equation (\ref{sparsity}). 
As shown in Table \ref{tab:ablation sparsity choice}, we fix $M =0.55$ with stage divider at steps 450 and 900 and keeping a total sparsity of all stages at 0.3 with sampling steps at 20. The results indicate that increasing the sparsity in Stage 2 and 3 leads to a noticeable degradation in performance. In contrast, increasing the sparsity in Stage 1 has minimal impact on FID and IS, suggesting that early-stage pruning is less detrimental to overall generation quality.
\begin{table}[t]
  \centering
  \caption{Ablation study on the sparsity choice strategy. }
  \label{tab:ablation sparsity choice}
  \vspace{-1em}
  
  \resizebox{0.45\textwidth}{!}{
  \begin{tabular}{l|ccc|cc}
    \toprule
    \multirow{2}{*}{\textbf{Strategy}} & 
    \multicolumn{3}{c|}{\textbf{Sparsity}} &
    \multirow{2}{*}{\textbf{IS}\(\uparrow\)} &
    \multirow{2}{*}{\textbf{FID}\(\downarrow\)} \\
    & \textbf{Stage 1}& 
    \textbf{Stage 2} &
    \textbf{Stage 3} &
    \\
    \midrule
    Uniform  & 0.3 &	0.3& 0.3& 80.6  & 33.21  \\
    Sparser Stage 1 & 0.9 & 0.04 & 0.1  & 245.8 & 3.53 \\
    Sparser Stage 2 & 0.6 & 0.15 & 0.1  & 176.4 & 10.20 \\
    Non SNR refined & 0.6 & 0.04 & 0.4 & 239.5 & 4.18 \\
    \rowcolor{cyan!10} {\em MosaicDiff} (Ours) & 0.6 & 0.04 & 0.1 & \textbf{266.1} & \textbf{3.22} \\
    \bottomrule
  \end{tabular}}
  \vspace{-1.5em}
\end{table}

\subsection{Analysis}

\begin{figure}[t]
    \centering
    \includegraphics[width=0.7\linewidth]{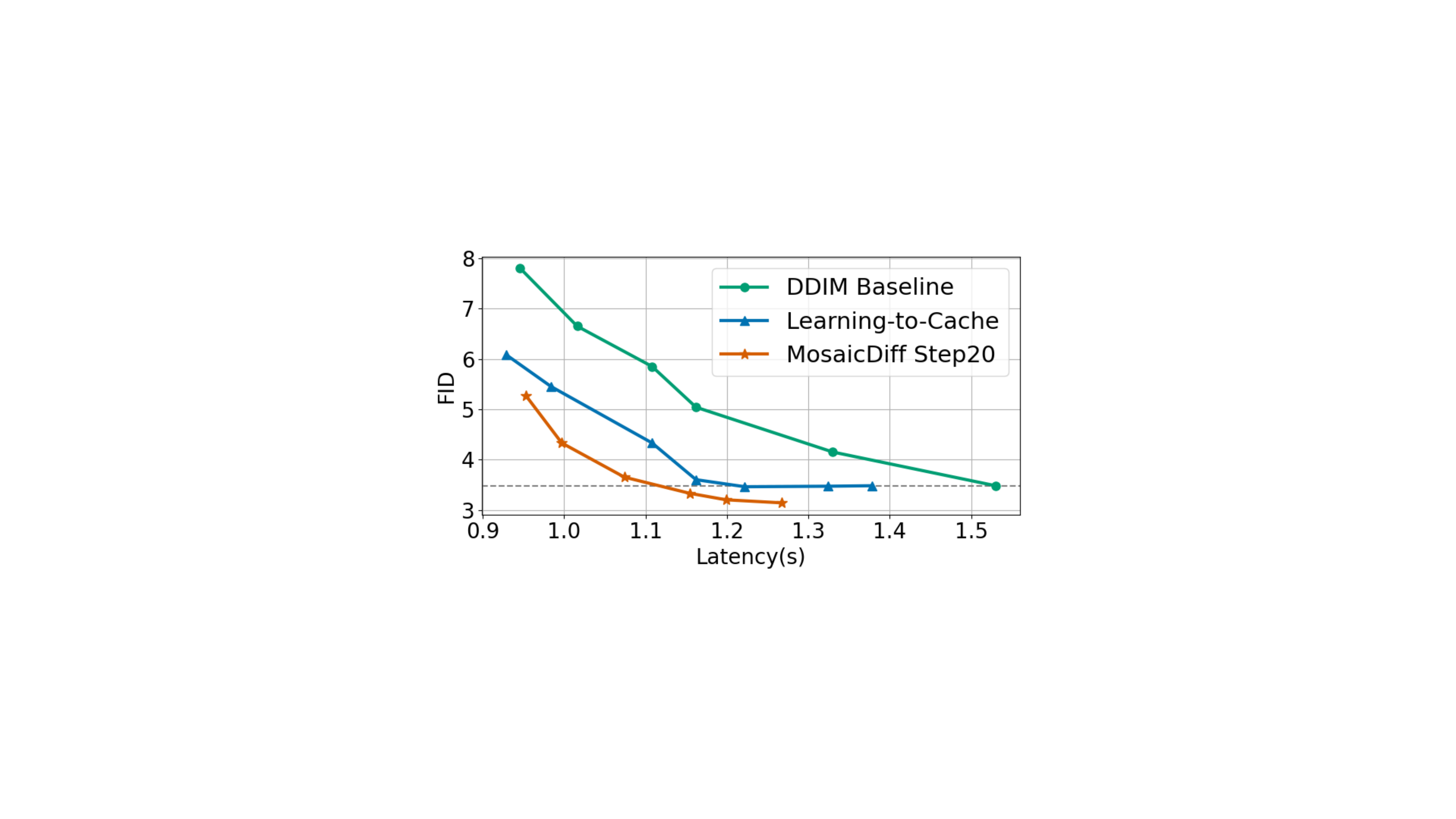}
    \vspace{-1em}
    \caption{Accuracy-Efficiency Tradeoff for DiT-XL/2 with 20 sampling steps. The dashed line represents the FID of the vanilla DDIM baseline at 20 sampling steps.}
    \label{fig:tradeoff}
    \vspace{-1em}
\end{figure}

\noindent \textbf{Accuracy-Efficiency Tradeoff.} Figure \ref{fig:tradeoff} illustrates the trade-off between FID and latency, providing a more comprehensive comparison of our method against the state-of-the-art caching algorithm and the vanilla model. Our approach consistently achieves superior performance across different sparsity levels. Specifically, for a given latency, {\bf{\em MosaicDiff}} achieves lower FID scores compared to both baselines. Conversely, for a target FID, {\bf{\em MosaicDiff}} consistently requires less latency, demonstrating its efficiency. The results highlight that {\bf{\em MosaicDiff}} maintains high generation quality while reducing computational cost, even under aggressive sparsity configurations, demonstrating its effectiveness in balancing accuracy and efficiency.

\noindent \textbf{Compatibility with Existing Acceleration Methods.} We demonstrate the compatibility of our approach with various existing acceleration strategies, including caching methods, fast samplers and step-distilled models. As shown in Table~\ref{tab:vertical}, combining our pruning technique with DeepCache substantially reduces MACs and improves FID scores. Integrating with the fast sampler DPM-Solver++ \cite{lu2022dpm++} further enhances image quality and reduces latency. Additionally, our method seamlessly applies to step-distilled models like SDXL-Turbo \cite{sauer2024adversarial}, simultaneously lowering computational cost and improving generation quality (see Appendix for more details). These results highlight the versatility and broad applicability of {\bf{\em MosaicDiff}} in conjunction with complementary acceleration methods.

\begin{table}[t]
  \centering
  \caption{Performance of {\bf{\em MosaicDiff}} at sparsity 0.15 when combined with existing acceleration strategies on SDXL.}
  \vspace{-0.6em}
  
  \label{tab:vertical}
  \resizebox{0.37\textwidth}{!}{
  \begin{tabular}{l|c|c|c}
    \toprule
    \textbf{Method} & 
    \textbf{Steps} & 
    \textbf{MACs (T)} & 
    \textbf{FID} \(\downarrow\) \\
    \midrule
    DeepCache-N=3  & 50 & 59.66 & 24.61\\
    Ours + DeepCache-N=3 & 50 & 50.71& \textbf{23.88}\\
    \midrule
    DPM-solver++  & 20 & 67.80 & 24.96\\
    Ours + DPM-solver++ & 20 & 57.63&\textbf{24.01}\\
    \midrule
    SDXL-Turbo  & 4 & 13.56 & 30.93\\
    Ours + SDXL-Turbo & 4 & 11.53&\textbf{30.08}\\
    \bottomrule
  \end{tabular}}
  \vspace{-0.7em}
\end{table}

\begin{table}[ht]
  \centering
  \caption{Runtime of {\bf{\em MosaicDiff}} and other compression methods.}
  \vspace{-1em}
  \label{tab:complexity}
  \resizebox{0.4\textwidth}{!}{
  \begin{tabular}{l|c|c|c|c}
    \toprule
    \textbf{Model} & \textbf{Resolution} &\textbf{Method} & \textbf{Hardware} & \textbf{GPU hours} \(\downarrow\)\\
    \midrule
    \multirow{2}{*}{DiT-XL/2}& \multirow{2}{*}{$256\times256$} & L2C & RTX 4090 &16.3\\
    &  & \cellcolor{cyan!10}{\em MosaicDiff} & \cellcolor{cyan!10}RTX 4090 & \cellcolor{cyan!10}\textbf{0.5}\\
    \midrule
    \multirow{2}{*}{SD-1.5}& \multirow{2}{*}{$512\times512$} & DiP-GO & MI250 &2.5\\
    &  & \cellcolor{cyan!10}{\em MosaicDiff} & \cellcolor{cyan!10}RTX 4090 & \cellcolor{cyan!10}\textbf{0.8}\\
    \midrule
    SDXL & $1024\times1024$ & {\em MosaicDiff} & RTX 4090 & 6.0\\
    \bottomrule
  \end{tabular}
  }
  \vspace{-1.5em}
\end{table}

\noindent \textbf{Runtime of Compression.} As a training-free approach, the primary computational cost of {\bf{\em MosaicDiff}} is Hessian matrix calculation. Table~\ref{tab:complexity} shows that our method prunes a full three-stage DiT-XL/2 (675M parameters) within 30 minutes using a single RTX 4090 GPU. In contrast, L2C requires over 16 hours of router training on identical hardware. DiP-GO takes approximately 2.5 hours to train a pruner for SD-1.5 (865M parameters) on an AMD MI250 GPU \cite{zhu2024dip}, whereas our method prunes it in just 0.8 hours on one RTX 4090. Furthermore, {\bf{\em MosaicDiff}} compresses the significantly larger SDXL model (2.6B parameters) at higher resolution in merely 6 hours. These results highlight the practicality, speed and ease-of-integration advantages of {\bf{\em MosaicDiff}} compared to training-based or fine-tuning-dependent methods for accelerating diffusion models.
\vspace{-0.3em}
\section{Conclusion}

We have presented {\bf{\em MosaicDiff}}, a training-free structural pruning framework applicable to both transformer-based and U-Net-based diffusion models. By leveraging the varying learning speeds across the denoising process, our approach assigns varying sparsity levels to different sampling stages. Extensive experiments demonstrate that the proposed method consistently outperforms state-of-the-art compression methods, confirming its effectiveness in accelerating diffusion models without sacrificing generation quality. While this approach introduces slightly more memory overhead due to the need to store multiple pruned sub-networks corresponding to different sampling stages, we leave exploring and optimizing this for future work.

{
    \small
    \bibliographystyle{ieeenat_fullname}
    \bibliography{main}

\begin{thebibliography}{46}
\providecommand{\natexlab}[1]{#1}
\providecommand{\url}[1]{\texttt{#1}}
\expandafter\ifx\csname urlstyle\endcsname\relax
  \providecommand{\doi}[1]{doi: #1}\else
  \providecommand{\doi}{doi: \begingroup \urlstyle{rm}\Url}\fi

\bibitem[Choi et~al.(2022)Choi, Lee, Shin, Kim, Kim, and Yoon]{choi2022perception}
Jooyoung Choi, Jungbeom Lee, Chaehun Shin, Sungwon Kim, Hyunwoo Kim, and Sungroh Yoon.
\newblock Perception prioritized training of diffusion models.
\newblock In \emph{Proceedings of the IEEE/CVF conference on computer vision and pattern recognition}, pages 11472--11481, 2022.

\bibitem[Deng et~al.(2009)Deng, Dong, Socher, Li, Li, and Fei-Fei]{deng2009imagenet}
Jia Deng, Wei Dong, Richard Socher, Li-Jia Li, Kai Li, and Li Fei-Fei.
\newblock Imagenet: A large-scale hierarchical image database.
\newblock In \emph{2009 IEEE conference on computer vision and pattern recognition}, pages 248--255. Ieee, 2009.

\bibitem[Dhariwal and Nichol(2021)]{dhariwal2021diffusion}
Prafulla Dhariwal and Alexander Nichol.
\newblock Diffusion models beat gans on image synthesis.
\newblock \emph{Advances in Neural Information Processing Systems}, 34:\penalty0 8780--8794, 2021.

\bibitem[Fang et~al.(2023{\natexlab{a}})Fang, Ma, and Wang]{fang2023pruning}
Gongfan Fang, Xinyin Ma, and Xinchao Wang.
\newblock Structural pruning for diffusion models.
\newblock In \emph{Advances in Neural Information Processing Systems}, pages 16716--16728. Curran Associates, Inc., 2023{\natexlab{a}}.

\bibitem[Fang et~al.(2023{\natexlab{b}})Fang, Ma, and Wang]{fang2023structural}
Gongfan Fang, Xinyin Ma, and Xinchao Wang.
\newblock Structural pruning for diffusion models.
\newblock In \emph{Advances in Neural Information Processing Systems}, 2023{\natexlab{b}}.

\bibitem[Frantar and Alistarh(2023)]{frantar2023sparsegpt}
Elias Frantar and Dan Alistarh.
\newblock Sparsegpt: Massive language models can be accurately pruned in one-shot.
\newblock In \emph{International conference on machine learning}, pages 10323--10337. PMLR, 2023.

\bibitem[Hassibi et~al.(1993)Hassibi, Stork, and Wolff]{298572}
B. Hassibi, D.G. Stork, and G.J. Wolff.
\newblock Optimal brain surgeon and general network pruning.
\newblock In \emph{IEEE International Conference on Neural Networks}, pages 293--299 vol.1, 1993.

\bibitem[He et~al.(2023)He, Liu, Liu, Wu, Zhou, and Zhuang]{he2023ptqd}
Yefei He, Luping Liu, Jing Liu, Weijia Wu, Hong Zhou, and Bohan Zhuang.
\newblock Ptqd: Accurate post-training quantization for diffusion models.
\newblock \emph{arXiv preprint arXiv:2305.10657}, 2023.

\bibitem[Heusel et~al.(2017)Heusel, Ramsauer, Unterthiner, Nessler, and Hochreiter]{heusel2017gans}
Martin Heusel, Hubert Ramsauer, Thomas Unterthiner, Bernhard Nessler, and Sepp Hochreiter.
\newblock Gans trained by a two time-scale update rule converge to a local nash equilibrium.
\newblock \emph{Advances in neural information processing systems}, 30, 2017.

\bibitem[Ho and Salimans(2022)]{ho2022classifier}
Jonathan Ho and Tim Salimans.
\newblock Classifier-free diffusion guidance.
\newblock \emph{arXiv preprint arXiv:2207.12598}, 2022.

\bibitem[Ho et~al.(2020)Ho, Jain, and Abbeel]{ho2020denoising}
Jonathan Ho, Ajay Jain, and Pieter Abbeel.
\newblock Denoising diffusion probabilistic models.
\newblock \emph{Advances in Neural Information Processing Systems}, 33:\penalty0 6840--6851, 2020.

\bibitem[Kim et~al.(2023)Kim, Song, Castells, and Choi]{kim2023architectural}
Bo-Kyeong Kim, Hyoung-Kyu Song, Thibault Castells, and Shinkook Choi.
\newblock On architectural compression of text-to-image diffusion models.
\newblock 2023.

\bibitem[Kim et~al.(2024)Kim, Song, Castells, and Choi]{kim2024bk}
Bo-Kyeong Kim, Hyoung-Kyu Song, Thibault Castells, and Shinkook Choi.
\newblock Bk-sdm: A lightweight, fast, and cheap version of stable diffusion.
\newblock In \emph{European Conference on Computer Vision}, pages 381--399. Springer, 2024.

\bibitem[Kurti{\'c} et~al.(2023)Kurti{\'c}, Frantar, and Alistarh]{kurtic2023ziplm}
Eldar Kurti{\'c}, Elias Frantar, and Dan Alistarh.
\newblock Ziplm: Inference-aware structured pruning of language models.
\newblock \emph{Advances in Neural Information Processing Systems}, 36:\penalty0 65597--65617, 2023.

\bibitem[Li* et~al.(2025)Li*, Lin*, Zhang*, Cai, Li, Guo, Xie, Meng, Zhu, and Han]{li2024svdquant}
Muyang Li*, Yujun Lin*, Zhekai Zhang*, Tianle Cai, Xiuyu Li, Junxian Guo, Enze Xie, Chenlin Meng, Jun-Yan Zhu, and Song Han.
\newblock Svdquant: Absorbing outliers by low-rank components for 4-bit diffusion models.
\newblock In \emph{The Thirteenth International Conference on Learning Representations}, 2025.

\bibitem[Li et~al.(2023)Li, Lian, Liu, Yang, Dong, Kang, Zhang, and Keutzer]{li2023q}
Xiuyu Li, Long Lian, Yijiang Liu, Huanrui Yang, Zhen Dong, Daniel Kang, Shanghang Zhang, and Kurt Keutzer.
\newblock Q-diffusion: Quantizing diffusion models.
\newblock \emph{arXiv preprint arXiv:2302.04304}, 2023.

\bibitem[Lin et~al.(2024)Lin, Liu, Li, and Yang]{lin2024common}
Shanchuan Lin, Bingchen Liu, Jiashi Li, and Xiao Yang.
\newblock Common diffusion noise schedules and sample steps are flawed.
\newblock In \emph{Proceedings of the IEEE/CVF winter conference on applications of computer vision}, pages 5404--5411, 2024.

\bibitem[Lin et~al.(2014)Lin, Maire, Belongie, Hays, Perona, Ramanan, Doll{\'a}r, and Zitnick]{lin2014microsoft}
Tsung-Yi Lin, Michael Maire, Serge Belongie, James Hays, Pietro Perona, Deva Ramanan, Piotr Doll{\'a}r, and C~Lawrence Zitnick.
\newblock Microsoft coco: Common objects in context.
\newblock In \emph{Computer Vision--ECCV 2014: 13th European Conference, Zurich, Switzerland, September 6-12, 2014, Proceedings, Part V 13}, pages 740--755. Springer, 2014.

\bibitem[Lu et~al.(2022{\natexlab{a}})Lu, Zhou, Bao, Chen, Li, and Zhu]{lu2022dpm}
Cheng Lu, Yuhao Zhou, Fan Bao, Jianfei Chen, Chongxuan Li, and Jun Zhu.
\newblock Dpm-solver: A fast ode solver for diffusion probabilistic model sampling in around 10 steps.
\newblock \emph{Advances in Neural Information Processing Systems}, 35:\penalty0 5775--5787, 2022{\natexlab{a}}.

\bibitem[Lu et~al.(2022{\natexlab{b}})Lu, Zhou, Bao, Chen, Li, and Zhu]{lu2022dpm++}
Cheng Lu, Yuhao Zhou, Fan Bao, Jianfei Chen, Chongxuan Li, and Jun Zhu.
\newblock Dpm-solver++: Fast solver for guided sampling of diffusion probabilistic models.
\newblock \emph{arXiv preprint arXiv:2211.01095}, 2022{\natexlab{b}}.

\bibitem[Luo et~al.(2023)Luo, Tan, Huang, Li, and Zhao]{luo2023latent}
Simian Luo, Yiqin Tan, Longbo Huang, Jian Li, and Hang Zhao.
\newblock Latent consistency models: Synthesizing high-resolution images with few-step inference.
\newblock \emph{arXiv preprint arXiv:2310.04378}, 2023.

\bibitem[Lyu et~al.(2022)Lyu, Xu, Yang, Lin, and Dai]{lyu2022accelerating}
Zhaoyang Lyu, Xudong Xu, Ceyuan Yang, Dahua Lin, and Bo Dai.
\newblock Accelerating diffusion models via early stop of the diffusion process.
\newblock \emph{arXiv preprint arXiv:2205.12524}, 2022.

\bibitem[Ma et~al.(2024{\natexlab{a}})Ma, Fang, Bi~Mi, and Wang]{ma2024learning}
Xinyin Ma, Gongfan Fang, Michael Bi~Mi, and Xinchao Wang.
\newblock Learning-to-cache: Accelerating diffusion transformer via layer caching.
\newblock \emph{Advances in Neural Information Processing Systems}, 37:\penalty0 133282--133304, 2024{\natexlab{a}}.

\bibitem[Ma et~al.(2024{\natexlab{b}})Ma, Fang, and Wang]{ma2024deepcache}
Xinyin Ma, Gongfan Fang, and Xinchao Wang.
\newblock Deepcache: Accelerating diffusion models for free.
\newblock In \emph{Proceedings of the IEEE/CVF conference on computer vision and pattern recognition}, pages 15762--15772, 2024{\natexlab{b}}.

\bibitem[Meng et~al.(2022)Meng, Gao, Kingma, Ermon, Ho, and Salimans]{meng2022distillation}
Chenlin Meng, Ruiqi Gao, Diederik~P Kingma, Stefano Ermon, Jonathan Ho, and Tim Salimans.
\newblock On distillation of guided diffusion models.
\newblock \emph{arXiv preprint arXiv:2210.03142}, 2022.

\bibitem[Meng et~al.(2023)Meng, Rombach, Gao, Kingma, Ermon, Ho, and Salimans]{meng2023distillation}
Chenlin Meng, Robin Rombach, Ruiqi Gao, Diederik Kingma, Stefano Ermon, Jonathan Ho, and Tim Salimans.
\newblock On distillation of guided diffusion models.
\newblock In \emph{Proceedings of the IEEE/CVF Conference on Computer Vision and Pattern Recognition}, pages 14297--14306, 2023.

\bibitem[Nichol and Dhariwal(2021)]{nichol2021improveddenoisingdiffusionprobabilistic}
Alex Nichol and Prafulla Dhariwal.
\newblock Improved denoising diffusion probabilistic models, 2021.

\bibitem[Pan et~al.(2024)Pan, Zhuang, Huang, Nie, Yu, Xiao, Cai, and Anandkumar]{pan2024tstitch}
Zizheng Pan, Bohan Zhuang, De-An Huang, Weili Nie, Zhiding Yu, Chaowei Xiao, Jianfei Cai, and Anima Anandkumar.
\newblock T-stitch: Accelerating sampling in pre-trained diffusion models with trajectory stitching.
\newblock \emph{arXiv}, 2024.

\bibitem[Peebles and Xie(2023)]{peebles2023scalable}
William Peebles and Saining Xie.
\newblock Scalable diffusion models with transformers.
\newblock In \emph{Proceedings of the IEEE/CVF international conference on computer vision}, pages 4195--4205, 2023.

\bibitem[Podell et~al.()Podell, English, Lacey, Blattmann, Dockhorn, M{\"u}ller, Penna, and Rombach]{podellsdxl}
Dustin Podell, Zion English, Kyle Lacey, Andreas Blattmann, Tim Dockhorn, Jonas M{\"u}ller, Joe Penna, and Robin Rombach.
\newblock Sdxl: Improving latent diffusion models for high-resolution image synthesis.
\newblock In \emph{The Twelfth International Conference on Learning Representations}.

\bibitem[Poole et~al.(2022)Poole, Jain, Barron, and Mildenhall]{poole2022dreamfusion}
Ben Poole, Ajay Jain, Jonathan~T Barron, and Ben Mildenhall.
\newblock Dreamfusion: Text-to-3d using 2d diffusion.
\newblock \emph{arXiv preprint arXiv:2209.14988}, 2022.

\bibitem[Radford et~al.(2021)Radford, Kim, Hallacy, Ramesh, Goh, Agarwal, Sastry, Askell, Mishkin, Clark, et~al.]{radford2021learning}
Alec Radford, Jong~Wook Kim, Chris Hallacy, Aditya Ramesh, Gabriel Goh, Sandhini Agarwal, Girish Sastry, Amanda Askell, Pamela Mishkin, Jack Clark, et~al.
\newblock Learning transferable visual models from natural language supervision.
\newblock In \emph{International conference on machine learning}, pages 8748--8763. PmLR, 2021.

\bibitem[Rombach et~al.(2022)Rombach, Blattmann, Lorenz, Esser, and Ommer]{rombach2022high}
Robin Rombach, Andreas Blattmann, Dominik Lorenz, Patrick Esser, and Bj{\"o}rn Ommer.
\newblock High-resolution image synthesis with latent diffusion models.
\newblock In \emph{Proceedings of the IEEE/CVF Conference on Computer Vision and Pattern Recognition}, pages 10684--10695, 2022.

\bibitem[Ronneberger et~al.(2015)Ronneberger, Fischer, and Brox]{ronneberger2015u}
Olaf Ronneberger, Philipp Fischer, and Thomas Brox.
\newblock U-net: Convolutional networks for biomedical image segmentation.
\newblock In \emph{Medical Image Computing and Computer-Assisted Intervention--MICCAI 2015: 18th International Conference, Munich, Germany, October 5-9, 2015, Proceedings, Part III 18}, pages 234--241. Springer, 2015.

\bibitem[Salimans and Ho(2022)]{salimans2022progressive}
Tim Salimans and Jonathan Ho.
\newblock Progressive distillation for fast sampling of diffusion models.
\newblock \emph{arXiv preprint arXiv:2202.00512}, 2022.

\bibitem[Sauer et~al.(2024)Sauer, Lorenz, Blattmann, and Rombach]{sauer2024adversarial}
Axel Sauer, Dominik Lorenz, Andreas Blattmann, and Robin Rombach.
\newblock Adversarial diffusion distillation.
\newblock In \emph{European Conference on Computer Vision}, pages 87--103. Springer, 2024.

\bibitem[Shang et~al.(2023)Shang, Yuan, Xie, Wu, and Yan]{shang2023post}
Yuzhang Shang, Zhihang Yuan, Bin Xie, Bingzhe Wu, and Yan Yan.
\newblock Post-training quantization on diffusion models.
\newblock In \emph{Proceedings of the IEEE/CVF conference on computer vision and pattern recognition}, pages 1972--1981, 2023.

\bibitem[Song et~al.(2020)Song, Meng, and Ermon]{song2020denoising}
Jiaming Song, Chenlin Meng, and Stefano Ermon.
\newblock Denoising diffusion implicit models.
\newblock \emph{arXiv preprint arXiv:2010.02502}, 2020.

\bibitem[Song and Ermon(2019)]{song2019generative}
Yang Song and Stefano Ermon.
\newblock Generative modeling by estimating gradients of the data distribution.
\newblock \emph{Advances in neural information processing systems}, 32, 2019.

\bibitem[Song et~al.(2023)Song, Dhariwal, Chen, and Sutskever]{song2023consistency}
Yang Song, Prafulla Dhariwal, Mark Chen, and Ilya Sutskever.
\newblock Consistency models.
\newblock 2023.

\bibitem[Tang et~al.(2024)Tang, Wang, Ding, Liang, Li, and Xu]{tang2024adadiff}
Shengkun Tang, Yaqing Wang, Caiwen Ding, Yi Liang, Yao Li, and Dongkuan Xu.
\newblock Adadiff: Accelerating diffusion models through step-wise adaptive computation.
\newblock In \emph{European Conference on Computer Vision}, pages 73--90. Springer, 2024.

\bibitem[Tang et~al.(2025)Tang, Sieberling, Kurtic, Shen, and Alistarh]{tang2025darwinlm}
Shengkun Tang, Oliver Sieberling, Eldar Kurtic, Zhiqiang Shen, and Dan Alistarh.
\newblock Darwinlm: Evolutionary structured pruning of large language models.
\newblock \emph{arXiv preprint arXiv:2502.07780}, 2025.

\bibitem[Vaswani et~al.(2017)Vaswani, Shazeer, Parmar, Uszkoreit, Jones, Gomez, Kaiser, and Polosukhin]{vaswani2017attention}
Ashish Vaswani, Noam Shazeer, Niki Parmar, Jakob Uszkoreit, Llion Jones, Aidan~N Gomez, {\L}ukasz Kaiser, and Illia Polosukhin.
\newblock Attention is all you need.
\newblock \emph{Advances in neural information processing systems}, 30, 2017.

\bibitem[Wang et~al.(2004)Wang, Bovik, Sheikh, and Simoncelli]{1284395}
Zhou Wang, A.C. Bovik, H.R. Sheikh, and E.P. Simoncelli.
\newblock Image quality assessment: from error visibility to structural similarity.
\newblock \emph{IEEE Transactions on Image Processing}, 13\penalty0 (4):\penalty0 600--612, 2004.

\bibitem[Zhang et~al.(2024)Zhang, Jin, Dong, Khakzar, Torr, Stegmaier, and Kawaguchi]{zhang2024effortless}
Yang Zhang, Er Jin, Yanfei Dong, Ashkan Khakzar, Philip Torr, Johannes Stegmaier, and Kenji Kawaguchi.
\newblock Effortless efficiency: Low-cost pruning of diffusion models.
\newblock \emph{arXiv preprint arXiv:2412.02852}, 2024.

\bibitem[Zhu et~al.(2024)Zhu, Tang, Liu, Lu, Zheng, Peng, Li, Wang, Jiang, Tian, et~al.]{zhu2024dip}
Haowei Zhu, Dehua Tang, Ji Liu, Mingjie Lu, Jintu Zheng, Jinzhang Peng, Dong Li, Yu Wang, Fan Jiang, Lu Tian, et~al.
\newblock Dip-go: A diffusion pruner via few-step gradient optimization.
\newblock \emph{Advances in Neural Information Processing Systems}, 37:\penalty0 92581--92604, 2024.

\end{thebibliography}
}

\newpage
\appendix

\section*{\Large{Appendix}}

\section{Proof of Theorem 1.}
\label{app:proof}
\renewcommand{\qedsymbol}{}
\begin{proof}
Since we implement our {\bf{\em MosaicDiff}} on well-pretrained diffusion models, we can assume that the distribution of generated images $\hat{x}_0, \hat{x}_t$ is converge to training data $x_0, x_t$:
\begin{equation}
    p_\theta(\hat{x}_{0}) \rightarrow q(x_{0}), \quad p_\theta(\hat{x}_{t}) \rightarrow q(x_{t})
\end{equation}
Thus, we can get similar relation between $\hat{x}_0$ and $\hat{x}_t$:
\begin{equation}
    \hat{x}_t(\hat{x}_0,\,\epsilon) = \sqrt{\bar{\alpha}_t}\hat{x}_0 + \sqrt{1-\bar{\alpha}_t}\epsilon, \quad 
\, \epsilon \sim \mathcal{N}(\mathbf{0},\mathbf{I}).
\end{equation}
The expectation of MSE can be derived as:
\begin{equation}
    \begin{gathered}
\mathbb{E}\Bigl[\text{MSE}(t)\Bigr] = \frac{1}{d}\mathbb{E}\Bigl[\|\hat{x}_t-\hat{x}_0\|_{2}^2\Bigr] \\= \frac{1}{d}\mathbb{E}\Bigl[\|\sqrt{\bar{\alpha}_t}\hat{x}_0 + \sqrt{1-\bar{\alpha}_t}\epsilon-\hat{x}_0\|_{2}^2\Bigr]
\\=\frac{1}{d}\mathbb{E}\Bigl[\|(\sqrt{\bar{\alpha}_t}-1)\hat{x}_0 + \sqrt{1-\bar{\alpha}_t}\epsilon\|_{2}^2\Bigr]
\\=\frac{1}{d}\mathbb{E}\Bigl[(\sqrt{\bar{\alpha}_t}-1)^2\|\hat{x}_0\|_{2}^2 + ({1-\bar{\alpha}_t})\|\epsilon\|_{2}^2 \\+ 2(\sqrt{\bar{\alpha}_t}-1)({1-\bar{\alpha}_t})\hat{x}_0\epsilon\Bigr],
\end{gathered}
\end{equation}
Since $\mathbb{E}[\epsilon]= 0, \mathbb{E}[\|\epsilon\|_{2}^2]=\|\mathbf{I}\|_2^2$:
\begin{equation}
    \begin{gathered}
\mathbb{E}\Bigl[\text{MSE}(t)\Bigr] = \frac{1}{d}\Bigl[(1-\sqrt{\bar{\alpha}_t})^2\|\hat{x}_0\|_2^2 + (1 - \bar{\alpha}_t)\|\mathbf{I}\|_2^2\Bigr].
\end{gathered}
\end{equation}
Then, we can calculate gradient $\text{Grad}(t)(t > 0)$ as :
\begin{equation}
    \begin{gathered}
    \mathbb{E}\Bigl[\text{Grad}(t)\Bigr] = \mathbb{E}\Bigl[\text{MSE}(t)\Bigr] - \mathbb{E}\Bigl[\text{MSE}(t-1)\Bigr]
    \\=\frac{1}{d}\Bigl[((\bar{\alpha}_t - \bar{\alpha}_{t-1}) + 2(\sqrt{\bar{\alpha}_{t-1}}-\sqrt{\bar{\alpha}_t}))\|\hat{x}_0\|_2^2 \\- (\bar{\alpha}_t - \bar{\alpha}_{t-1}))\|\mathbf{I}\|_2^2\Bigr].
    \end{gathered}
\end{equation}
Define $\delta_t := \bar{\alpha}_t - \bar{\alpha}_{t-1}$. Thus,
\begin{equation}
    \begin{gathered}
    \mathbb{E}\Bigl[\text{Grad}(t)\Bigr] =\frac{1}{d}\Bigl[(\delta_t + 2(\sqrt{\bar{\alpha}_{t-1}}-\sqrt{\bar{\alpha}_t}))\|\hat{x}_0\|_2^2 - \delta_t\|\mathbf{I}\|_2^2\Bigr].
    \end{gathered}
\end{equation}

\end{proof}

\section{Additional Experimental Results.}

\subsection{Comparison with Small Models Trained from Scratch.}
We evaluate our pruned large-scale model in comparison to the smaller DiT-L/2 model \cite{peebles2023scalable, ma2024learning} trained from scratch, which contains 458 million parameters. Both models are sampled using DDIM with 50 and 20 steps. As shown in Table~\ref{tab:compare_small_model}, our pruned model consistently outperforms the smaller DiT-L/2 across all evaluated metrics, including FID, IS, and Precision, while requiring comparable or fewer MACs. This demonstrates that even after pruning, our large-scale model retains significant performance advantages over smaller models trained from scratch, highlighting the effectiveness of our approach in balancing efficiency and generative quality. 
\begin{table}[t]
  \centering
  \caption{Comparison between {\bf{\em MosaicDiff}} at sparsity 0.35 and the smaller DiT-L/2 model trained from scratch.}
  \vspace{-1em}
  \label{tab:compare_small_model}
  \resizebox{0.48\textwidth}{!}{
  \begin{tabular}{l|c|c|cccc}
    \toprule
    \textbf{Model} & 
    \textbf{Steps} & 
    \textbf{MACs(T)} &
    \textbf{IS} \(\uparrow\)&
    \textbf{FID} \(\downarrow\)&
    \textbf{Precision} \(\uparrow\)&
    \textbf{Recall} \(\uparrow\)\\
    \midrule
    DiT-L/2  &50 &	3.88	&167.6	& 4.82	&78.72&	54.66 \\
    Ours &50	&3.88	&\textbf{265.9}	&\textbf{2.26}	&\textbf{81.76}	&\textbf{57.21}  \\
    \midrule
    DiT-L/2 & 20	&1.55	&160.2	&6.45	&77.13	&53.65 \\
    Ours  & 20	&1.51	&\textbf{264.5}	&\textbf{3.33}	&\textbf{80.37}	&\textbf{53.72} \\
    \bottomrule
  \end{tabular}
  }
  \vspace{-1.5em}
\end{table}

\subsection{Sparsity Allocation}
\label{app:sparsity}
We provide the sparsity allocation for each stage and the corresponding performance, as shown in Table ~\ref{tab:sparsity dit} and ~\ref{tab:sparsity sdxl}.  These results demonstrate that our method maintains strong performance even at higher sparsity levels. In Table 11, our approach achieves an FID of 3.65 at 40\% sparsity, showing minimal degradation. While extreme pruning (50\% sparsity) impacts performance, our method remains effective by strategically allocating sparsity across stages. Table 12 further confirms this trend for SDXL, where our method achieves an FID of 23.79 at 20\% sparsity, maintaining competitive quality. Even at 30\% sparsity, the model still produces reasonable results. These findings highlight that our method successfully balances compression and generation quality, outperforming conventional pruning techniques, especially at higher sparsity levels.
 
\begin{table}[h]
  \centering
  \caption{Sparsity allocation of DiT when $M=0.55$, stage divided at Step $T =450$ and $T =900$.}
  \label{tab:sparsity dit}
  \vspace{-1em}

  \resizebox{0.35\textwidth}{!}{
  \begin{tabular}{l|ccc|c}
    \toprule
    \textbf{Sparsity} & 
    \textbf{Stage 1} & 
    \textbf{Stage 2} & 
    \textbf{Stage 3} &
    \textbf{FID}\\
    \midrule
    0.25  & 0.50 & 0.02 & 0.06 & 3.14\\
    0.30  & 0.60 & 0.04 & 0.10 & 3.20\\
    0.35  & 0.70 & 0.06 & 0.20 & 3.33\\
    0.40  & 0.80 & 0.08 & 0.30 & 3.65\\
    0.45  & 0.90 & 0.10 & 0.40 & 4.33\\
    0.50  & 0.90 & 0.15 & 0.40 & 5.27\\
    \bottomrule
  \end{tabular}}
\end{table}
\begin{table}[ht]
  \centering
  \caption{Sparsity allocation of SDXL when $M=0.55$, stage divided at Step $T =250$ and $T =900$.}
  \label{tab:sparsity sdxl}
  \vspace{-1em}

  \resizebox{0.35\textwidth}{!}{
  \begin{tabular}{l|ccc|c}
    \toprule
    \textbf{Sparsity} & 
    \textbf{Stage 1} & 
    \textbf{Stage 2} & 
    \textbf{Stage 3} &
    \textbf{FID} \\
    \midrule
    0.10  & 0.30 & 0.03 & 0.15 & 23.18\\
    0.15  & 0.40 & 0.04 & 0.20 & 23.73\\
    0.20  & 0.60 & 0.06 & 0.30 & 23.79\\
    0.30  & 0.80 & 0.08 & 0.40 & 28.37\\
    \bottomrule
  \end{tabular}}
\end{table}

\subsection{Usability on Step-distilled Models}
\begin{figure}[h]
    \centering
    \includegraphics[width=0.8\linewidth]{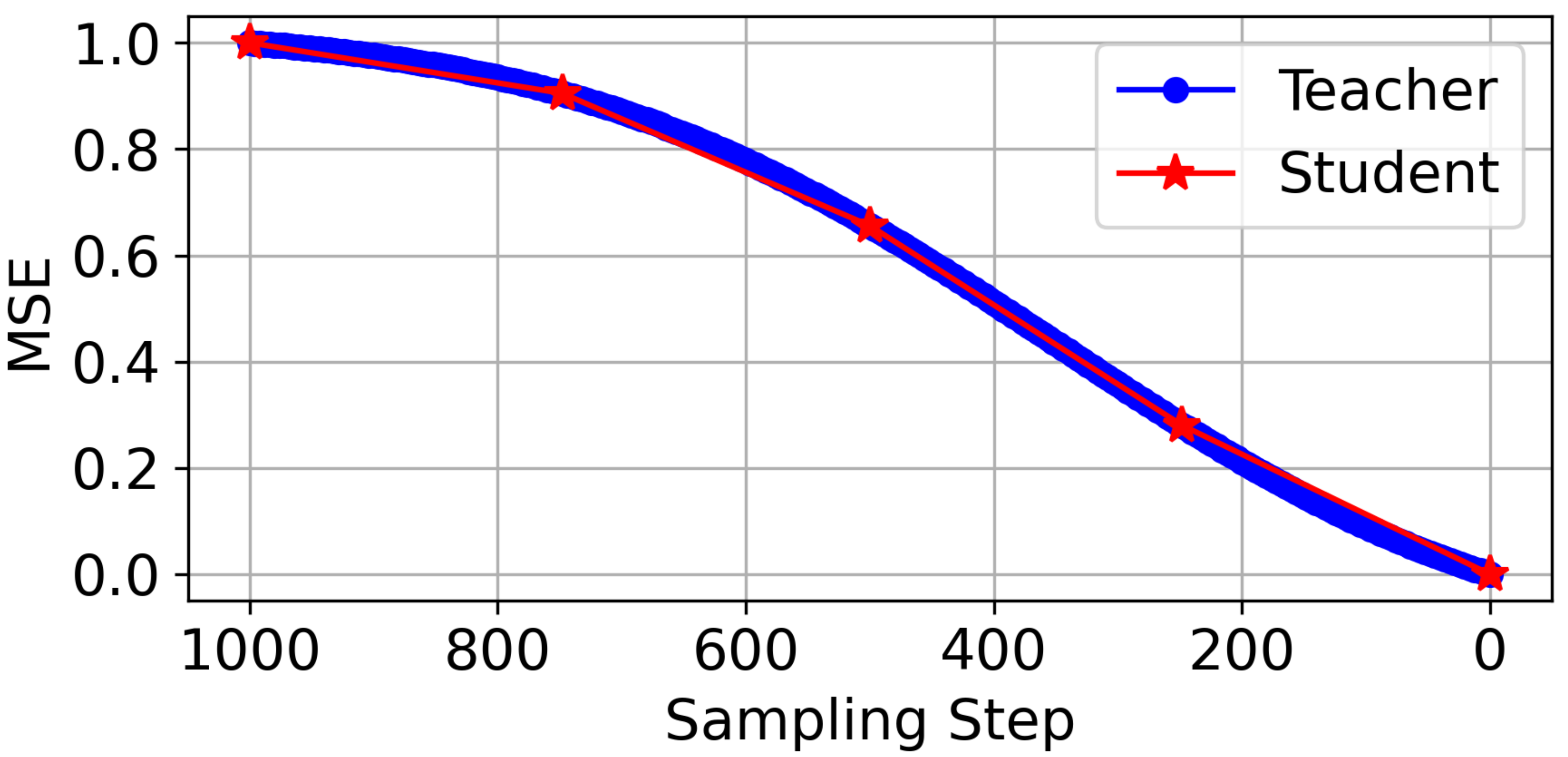}
    \vspace{-1em}
    \caption{Change in image MSE over sampling steps. Student SDXL-turbo aligns well with teacher SDXL.}
    \label{fig:student MSE}
    \vspace{-0.1in}
\end{figure}

{\bf{\em{MosaicDiff}}} is fully compatible with step-distilled models. We use SDXL-Turbo, a distilled variant of SDXL-Base-1.0, for evaluation. Experiments use 4 steps sampling. As in Table \ref{tab:step distill}, with 0.15 average sparsity, {\bf{\em{MosaicDiff}}} surpasses vanilla model and uniform pruning by FID margins of 0.85 and 0.69. In contrast, mismatched sparsity patterns degrade performance noticeably, validating our scoring strategy. We also show changes in image MSE over sampling steps, aligning well with the teacher (Figure \ref{fig:student MSE}).

\begin{table}[h]
  \centering
  \caption{Performance of {\bf{\em MosaicDiff}} on step-distilled model SDXL-turbo with 4 steps of sampling.}
  \label{tab:step distill}
  \vspace{-1em}
  
  \resizebox{0.44\textwidth}{!}{
  \begin{tabular}{c|cccc|c}
    \toprule
    \multirow{2}{*}{\textbf{Strategy}} & 
    \multicolumn{4}{c|}{\textbf{Sparsity}} &
    \multirow{2}{*}{\textbf{FID}\(\downarrow\)} \\
    & \textbf{Step 1}& 
    \textbf{Step 2} &
    \textbf{Step 3} &
    \textbf{Step 4} &
    \\
    \midrule
    Vanilla SDXL-turbo& 0 & 0 & 0 & 0 & 30.93 \\
    Uniform pruning & 0.15 & 0.15 & 0.15 & 0.15  & 30.77  \\
    Reverse \emph{MosaicDiff} & 0.05 & 0.1 & 0.3  & 0.15 & 31.86 \\
    \rowcolor{cyan!10}
    \emph{MosaicDiff} & 0.3 & 0.15 & 0.05 & 0.1 & \textbf{30.08} \\
    \bottomrule
  \end{tabular}}
  \vspace{-1.3em}
\end{table}

\subsection{Relationship between CFG and Sparsity} 
\begin{figure}[t]
    \centering
    \includegraphics[width=0.8\linewidth]{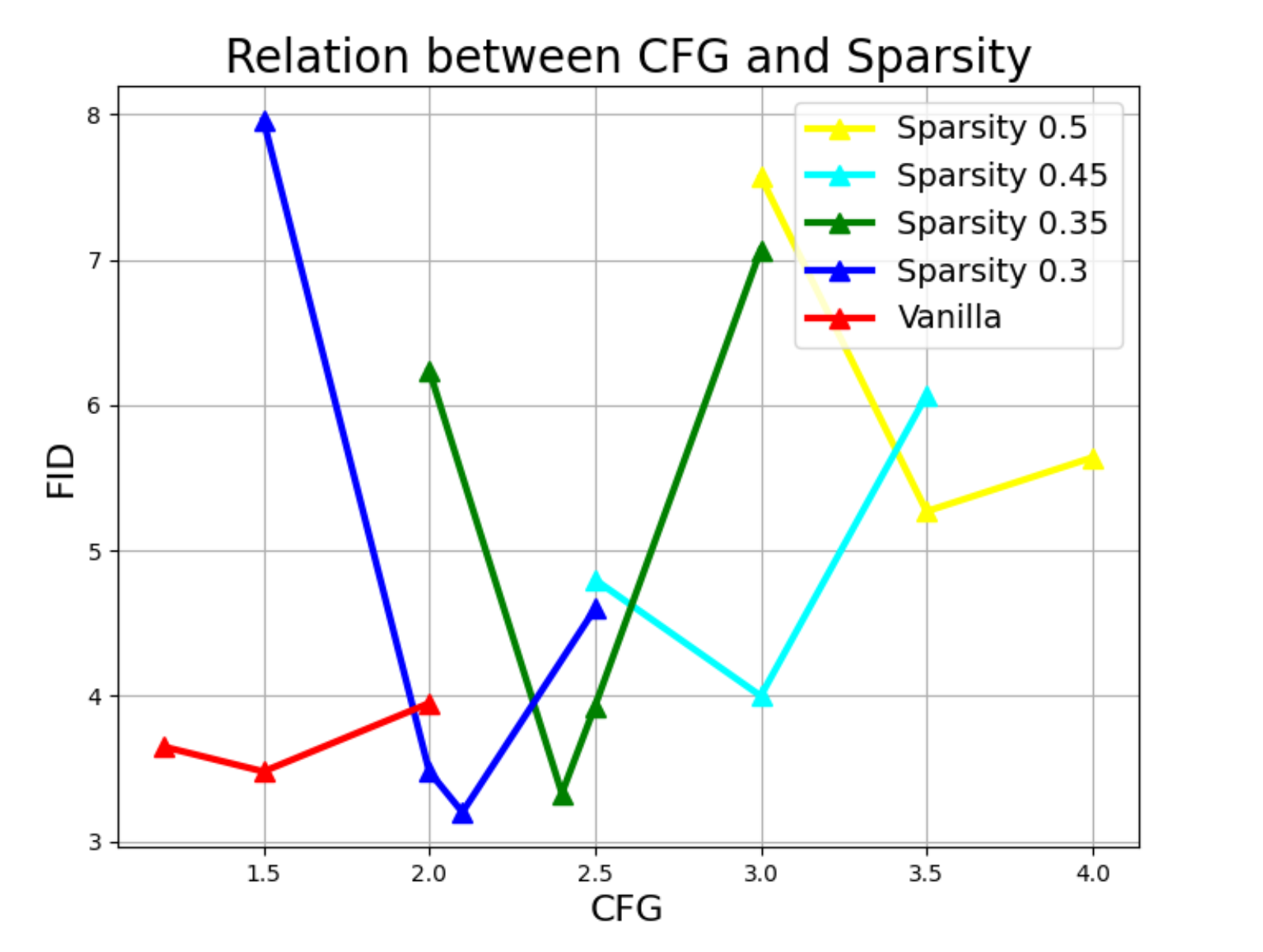}
    \vspace{-1em}
    \caption{Relationship between CFG and Sparsity.}
    \label{fig:CFG}
    \vspace{-0.1in}
\end{figure}
We observe that as pruning sparsity increases, the optimal CFG required to achieve the best FID also rises. Specifically, as illustrated in Figure~\ref{fig:CFG}, the optimal CFG value for the vanilla DiT-XL/2 model is approximately 1.5. At a pruning sparsity of 0.3, the optimal CFG increases to 2.1, and further increases to 3.5 at a sparsity level of 0.45. These results highlight a strong interplay between model compression and guidance strength.

\begin{figure}[t]
    \centering
    \includegraphics[width=1\linewidth]{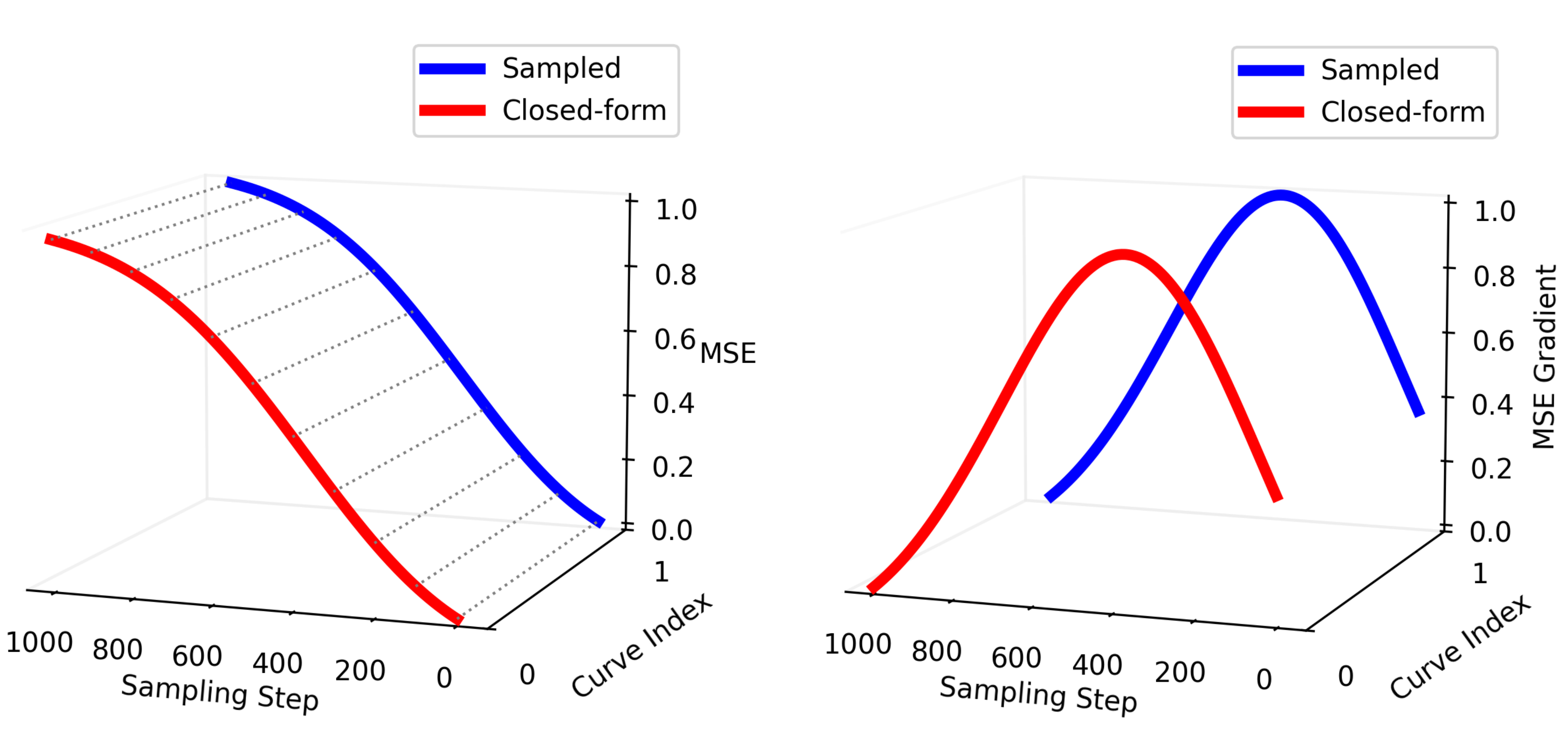}
    \vspace{-0.3in}
    \caption{MSE and gradient curves comparison under Scaled-Linear Schedule. \textit{Left}: MSE calculated from our closed-form approximation closely matches the sampled results. \textit{Right}: Gradients derived from our closed-form expression align with empirically sampled gradients. }
    \label{app:mse}
    \vspace{-0.1in}
\end{figure}

\begin{figure}[ht]
    \centering
    \begin{subfigure}{0.48\linewidth}
    \includegraphics[width=\linewidth]{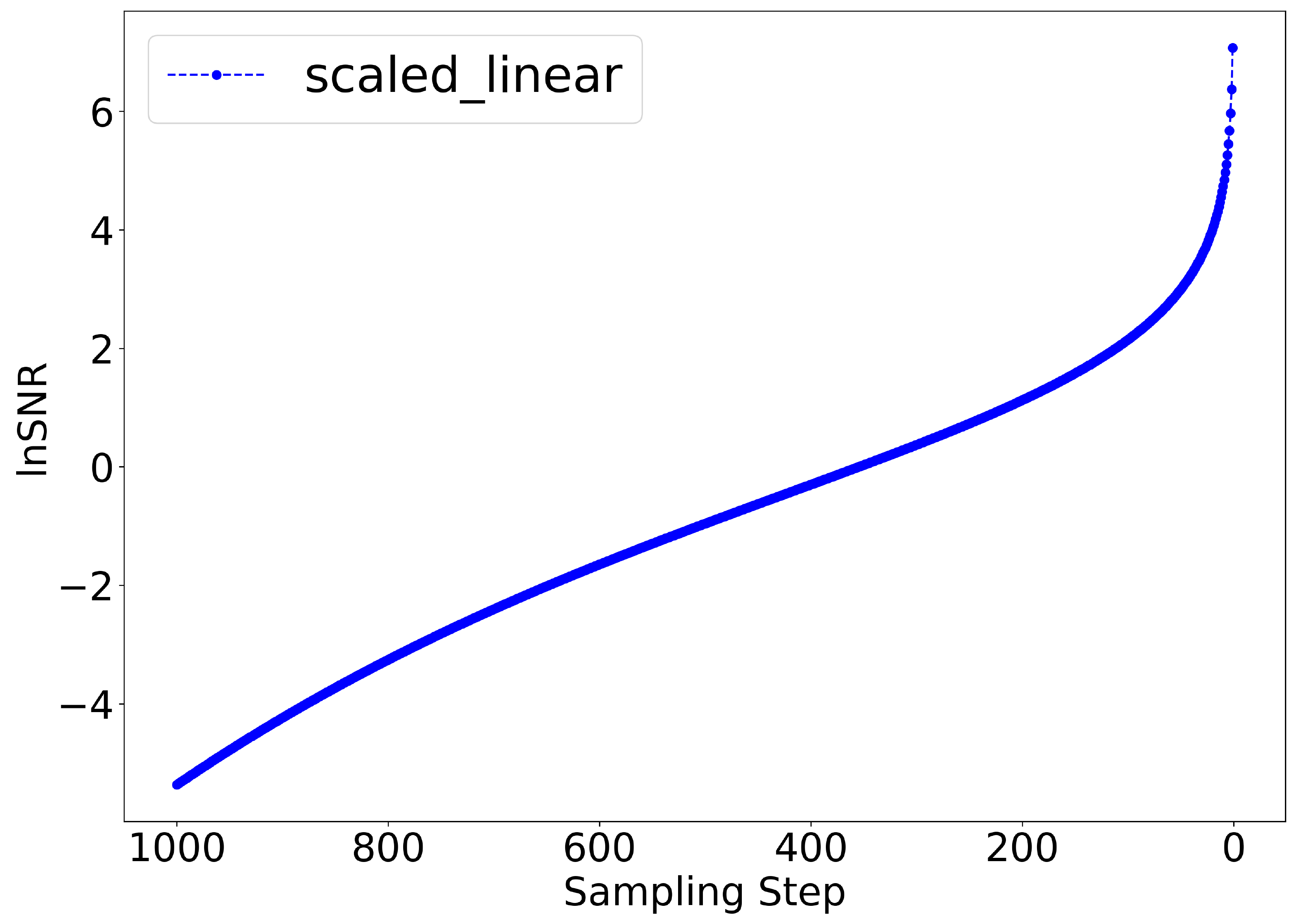}
    \caption{SNR trend of scale linear schedule.}
    \label{fig:SNR change app}
  \end{subfigure}
  \hfill
  \begin{subfigure}{0.48\linewidth}
    \includegraphics[width=\linewidth]{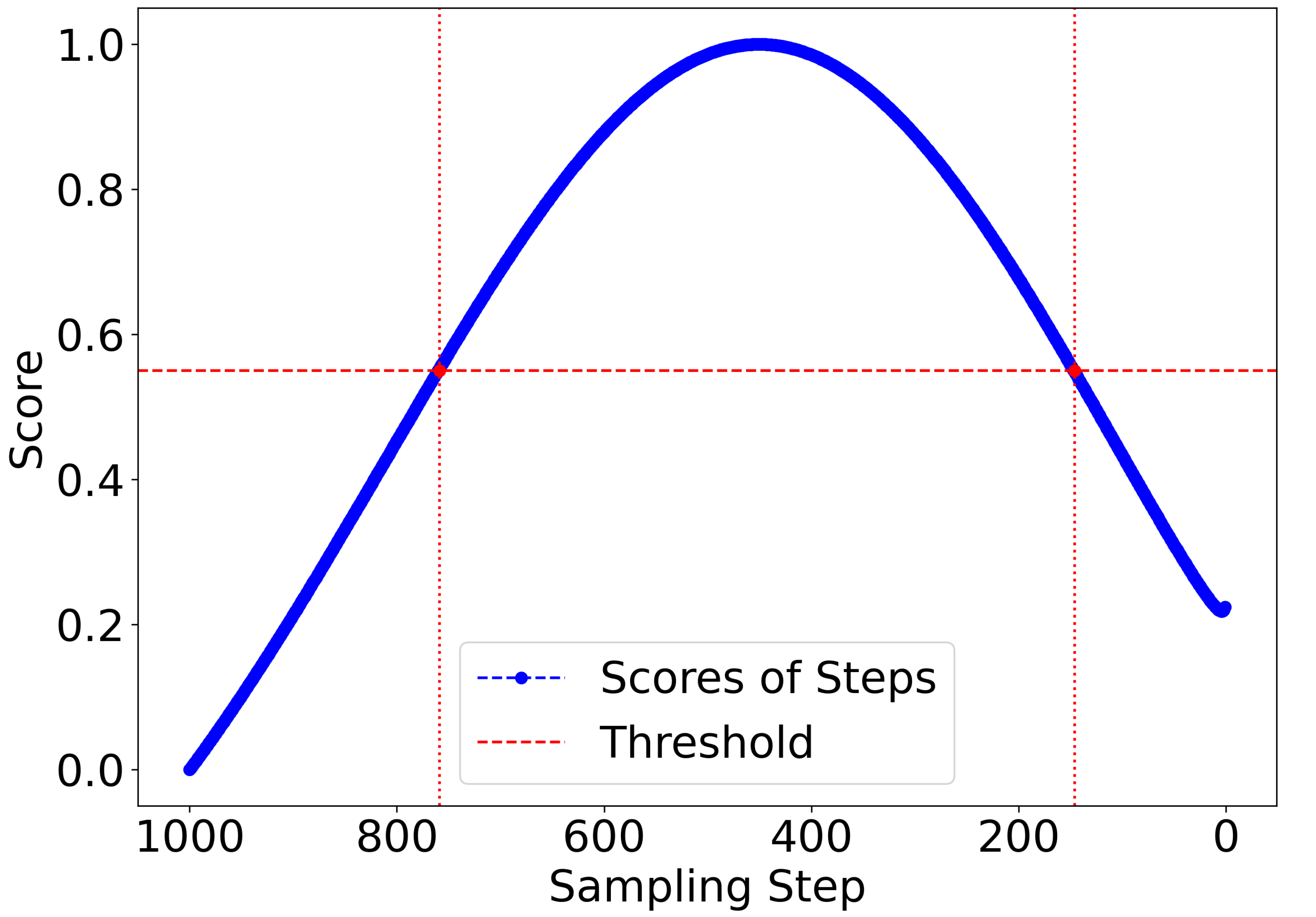}
    \caption{Final scores of sampling steps.}
    \label{app:Final scores}
  \end{subfigure}
  \caption{Influence of SNR on Final Scores. (a) Change in SNR across sampling steps, showing a sharp increase during the final steps. (b) Final scores computed combining SNR. A threshold of $M=0.55$ clearly divides the curve into three stages.}
  \vspace{-1em}
\end{figure}

\section{Additional Visualization of {\bf{\em MosaicDiff}}}
We provide the visualization of MSE and gradient on SDXL, as shown in Figure \ref{app:mse} and \ref{app:Final scores}. The results are similar as the figure we obtained in the method section.

\begin{figure}[t]
    \centering
    \includegraphics[width=1\linewidth]{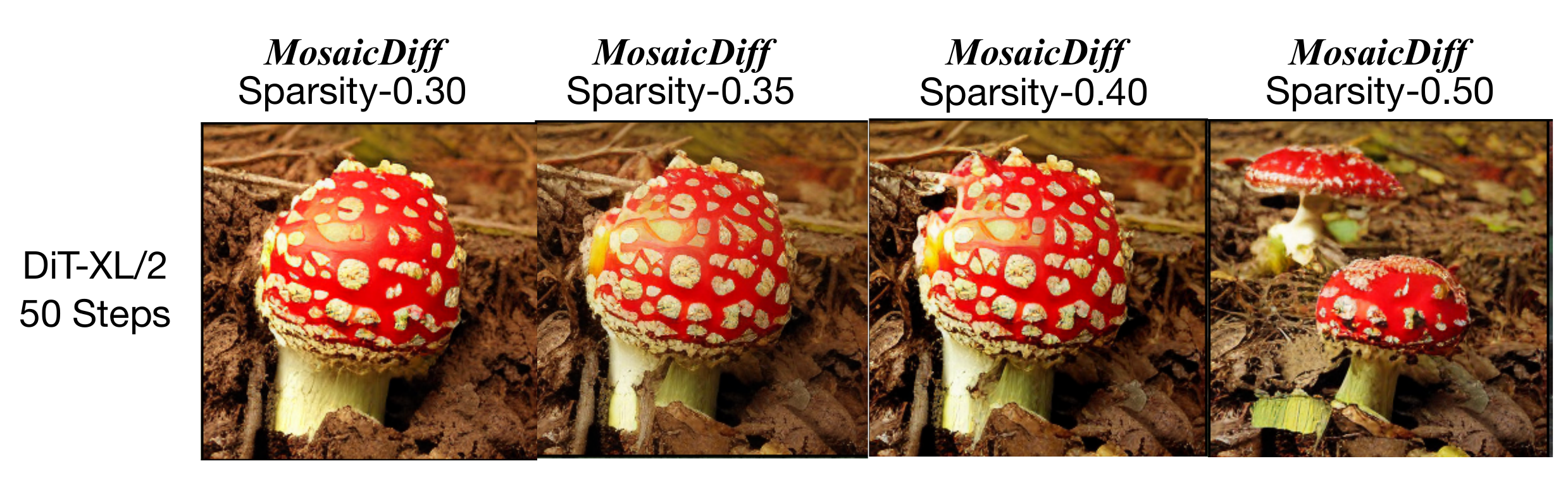}
    \vspace{-1em}
     \includegraphics[width=1\linewidth]{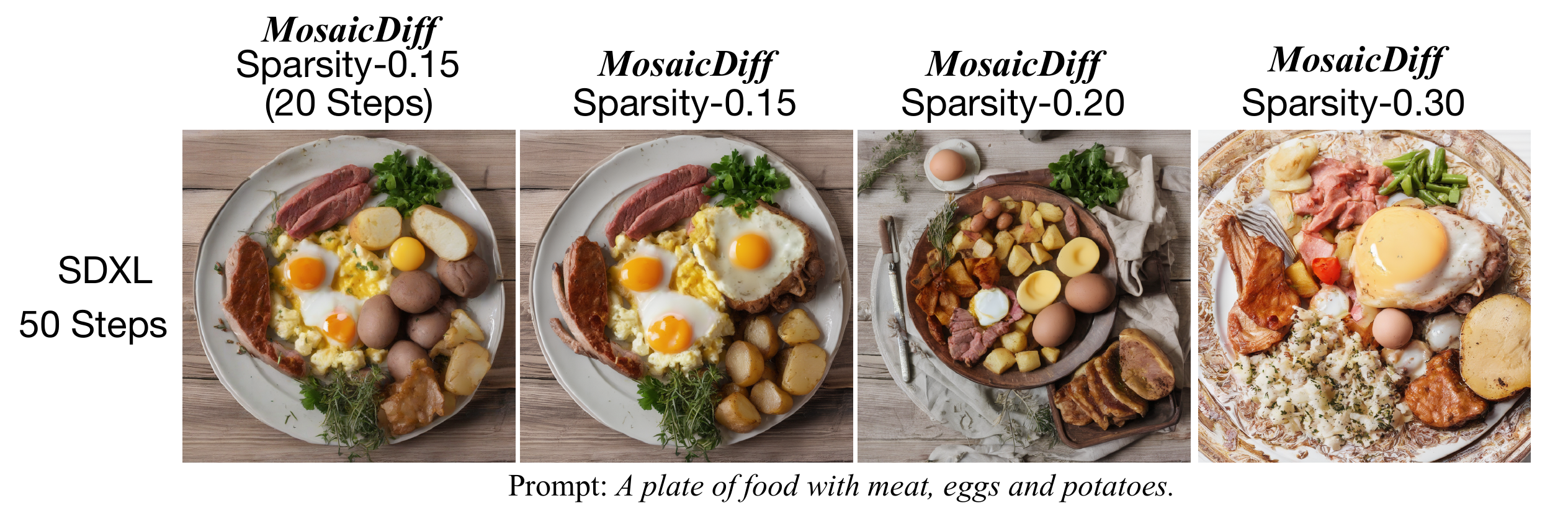}
     \caption{Generation Case from MosaicDiff on DiT and SDXL.}
    \label{app:visualization}
\end{figure}

Moreover, we add more visualization of images generated by MosaicDiff in Figure \ref{app:visualization}.

\end{document}